# A Unified Computational and Statistical Framework for Nonconvex Low-Rank Matrix Estimation

Lingxiao Wang[*‡]    and    Xiao Zhang[†‡]    and    Quanquan Gu[§]

## Abstract

We propose a unified framework for estimating low-rank matrices through nonconvex optimization based on gradient descent algorithm. Our framework is quite general and can be applied to both noisy and noiseless observations. In the general case with noisy observations, we show that our algorithm is guaranteed to linearly converge to the unknown low-rank matrix up to minimax optimal statistical error, provided an appropriate initial estimator. While in the generic noiseless setting, our algorithm converges to the unknown low-rank matrix at a linear rate and enables exact recovery with optimal sample complexity. In addition, we develop a new initialization algorithm to provide a desired initial estimator, which outperforms existing initialization algorithms for nonconvex low-rank matrix estimation. We illustrate the superiority of our framework through three examples: matrix regression, matrix completion, and one-bit matrix completion. We also corroborate our theory through extensive experiments on synthetic data.

## 1 Introduction

Low-rank matrix estimation has broad applications in many fields such as collaborative filtering (Srebro et al., 2004). Numerous efforts have been made in order to efficiently estimate the unknown low-rank matrix, among which the nuclear norm relaxation based methods (Srebro et al., 2004; Candès and Tao, 2010; Rohde et al., 2011; Recht et al., 2010; Recht, 2011; Negahban and Wainwright, 2011, 2012; Gui and Gu, 2015) are most popular. Although the nuclear norm based methods enjoy nice theoretical guarantees for recovering low-rank matrices, the computational complexities of these methods are very high. For example, to estimate a rank-$r$ matrix, most of these algorithms require to compute a rank-$r$ singular value decomposition per-iteration, which is computationally prohibitive for huge matrices. In order to get over such a computational barrier, many recent studies proposed to estimate the unknown low-rank matrix via matrix factorization, or more generally speaking, nonconvex optimization. Specifically, for a rank-$r$ matrix $\mathbf{X} \in \mathbb{R}^{d_1 \times d_2}$, it can be factorized as $\mathbf{X} = \mathbf{U}\mathbf{V}^\top$, where $\mathbf{U} \in \mathbb{R}^{d_1 \times r}$, $\mathbf{V} \in \mathbb{R}^{d_2 \times r}$, and such a reparametrization automatically enforces the low-rankness of the unknown matrix. While matrix factorization makes the optimization problem nonconvex, it can significantly improves the computational efficiency. A series of work (Jain et al., 2013; Zhao et al., 2015; Chen and Wainwright, 2015; Zheng and Lafferty, 2015; Tu et al., 2015; Bhojanapalli et al., 2015; Park et al., 2016a,b) has been carried out to analyze different nonconvex optimization algorithms for various low-rank matrix estimation problems.

---

[*]Department of Systems and Information Engineering, University of Virginia, Charlottesville, VA 22904, USA; e-mail: `lw4wr@virginia.edu`

[†]Department of Statistics, University of Virginia, Charlottesville, VA 22904, USA; e-mail:`xz7bc@virginia.edu`

[‡]Equal Contribution

[§]Department of Systems and Information Engineering, Department of Computer Science, University of Virginia, Charlottesville, VA 22904, USA; e-mail: `qg5w@virginia.edu`

In this paper, we propose a unified framework for nonconvex low-rank matrix estimation, which integrates both optimization-theoretic and statistical analyses. Our algorithm is applicable to both low-rank matrix estimation with noisy observations and that with noiseless observations. Instead of considering specific low-rank matrix estimation problems, we consider general ones, which correspond to optimizing a family of loss functions which satisfies restricted strongly convexity and smoothness conditions (Negahban et al., 2009). We establish the linear rate of convergence to the unknown low-rank matrix for our algorithm. In particular, for noisy observations, our algorithm achieves statistical error that matches the minimax lower bound (Negahban and Wainwright, 2012; Koltchinskii et al., 2011). While in the noiseless case, our algorithm enables exact recovery of the global optimum (i.e., unknown low-rank matrix) and achieves optimal sample complexity (Recht et al., 2010; Tu et al., 2015). Furthermore, we develop a new and generic initialization algorithm to provide suitable initial estimator. We prove that our initialization procedure relaxes the stringent requirement on condition number of the objective function, assumed in recent studies (Bhojanapalli et al., 2015; Park et al., 2016a,b), thereby resolving an open question in Bhojanapalli et al. (2015).

Furthermore, we apply our unified framework to specific problems, such as matrix sensing, matrix completion and one-bit matrix completion. We establish the linear convergence rates and optimal statistical error bounds of our method for each examples. We also demonstrate the superiority of our approach over the state-of-the art methods via thorough experiments.

**Notation.** We use capital symbols such as $\mathbf{A}$ to denote matrices and $[d]$ to denote $\{1, 2, \ldots, d\}$. The inner product between two matrices is denoted by $\langle \mathbf{A}, \mathbf{B} \rangle = \text{Tr}(\mathbf{A}^\top \mathbf{B})$. For any index set $\Omega \subseteq [d_1] \times [d_2]$, denote $\Omega_{i,*} = \{(i, j) \in \Omega \mid j \in [d_2]\}$, and $\Omega_{*,j} = \{(i, j) \in \Omega \mid i \in [d_1]\}$. For any matrix $\mathbf{A} \in \mathbb{R}^{d_1 \times d_2}$, we denote the $i$-th row and $j$-th column of $\mathbf{A}$ by $\mathbf{A}_{i,*}$ and $\mathbf{A}_{*,j}$, respectively. The $(i, j)$-th entry of $\mathbf{A}$ is denoted by $A_{ij}$. Denote the $\ell$-th largest singular value of $\mathbf{A}$ by $\sigma_\ell(\mathbf{A})$. For any matrix $\mathbf{A} \in \mathbb{R}^{(d_1+d_2) \times r}$, we use $\mathbf{A}_U$ and $\mathbf{A}_V$ to denote the top $d_1 \times r$ and bottom $d_2 \times r$ matrices of $\mathbf{A}$, respectively. Let $\mathbf{x} = [x_1, x_2, \cdots, x_d]^\top \in \mathbb{R}^d$ be a $d$-dimensional vector. For $0 < q < \infty$, denote the $\ell_q$ vector norm by $\|\mathbf{x}\|_q = (\Sigma_{i=1}^d |x_i|^q)^{1/q}$. As usual, let $\|\mathbf{A}\|_F$, $\|\mathbf{A}\|_2$ be the Frobenius norm and the spectral norm of matrix $\mathbf{A}$, respectively. The nuclear norm of $\mathbf{A}$ is defined as $\|\mathbf{A}\|_* = \sum_{i=1}^r \sigma_i(\mathbf{A})$, where $r$ denote the rank of $\mathbf{A}$, and the element-wise infinity norm of $\mathbf{A}$ is defined as $\|\mathbf{A}\|_\infty = \max_{i,j} |A_{ij}|$. Besides, we define the largest $\ell_2$ norm of its rows as $\|\mathbf{A}\|_{2,\infty} = \max_i \|\mathbf{A}_{i,*}\|_2$.

## 2 Related Work

In recent years, a surge of nonconvex optimization algorithms for estimating low-rank matrices have been established. For example, Jain et al. (2013) analyzed the convergence of alternating minimization approach for matrix sensing and matrix completion. Zhao et al. (2015) provided a more unified analysis by proving that, with a reasonable initial solution, a broad class of nonconvex optimization algorithms, including alternating minimization and gradient-type methods, can successfully recover the true low-rank matrix. However, they also required a stringent form of the restricted isometry property that is similar to Jain et al. (2013). Recently, Zheng and Lafferty (2015, 2016) analyzed the gradient descent based approach for matrix sensing and matrix completion.

They showed that their algorithm is guaranteed to converge linearly to the global optimum with an appropriate initial solution, and improves the alternating minimization algorithm in terms of both computational complexity and sample complexity. Tu et al. (2015) provided an improved analysis of matrix sensing via gradient descent, compared to Zheng and Lafferty (2015), through a more sophisticated initialization procedure and a refined restricted isometry assumption on the measurements.

The most related work to ours is Chen and Wainwright (2015); Bhojanapalli et al. (2015); Park et al. (2016b). In detail, Chen and Wainwright (2015) proposed a projected gradient descent framework to recover the positive semidefinite low-rank matrices. Although their work can be applied to a wide range of problems, the iteration complexity derived from their optimization framework is very high for many specific examples. Bhojanapalli et al. (2015) proposed a factorized gradient descent algorithm for nonconvex optimization over positive semidefinite matrices. They proved that, when the general empirical loss function is both strongly convex and smooth, their algorithm can recover the unknown low-rank matrix at a linear convergence rate. Built upon Bhojanapalli et al. (2015), Park et al. (2016b) derived the theoretical guarantees of the factorized gradient descent algorithm for rectangular matrix factorization problem under similar conditions. Nevertheless, their analyses (Bhojanapalli et al., 2015; Park et al., 2016b) are limited to the optimization perspective, and do not support the case with noisy observations. Our proposed framework, on one hand, simplifies the conditions of nonconvex low-rank matrix estimation to restricted strongly convexity and smoothness, and on the other hand, integrates both optimization-theoretic and statistical analyses. In fact, it achieves the best of both worlds, and provides a simple but powerful toolkit to analyze various low-rank matrix estimation problems. Furthermore, our proposed initialization algorithm relaxes the strict constraint on condition number of the objective function, which is imposed by Bhojanapalli et al. (2015); Park et al. (2016a,b), thereby resolving an open question in Bhojanapalli et al. (2015).

We also note that in order to get rid of the disadvantages of initialization procedure, Bhojanapalli et al. (2016); Park et al. (2016c) proved that for matrix sensing, all local minima of the nonconvex optimization based on matrix reparametrization are close to a global optimum under the restricted isometry property assumption. And for positive semidefinite matrix completion, Ge et al. (2016) proved a similar result. However, for general low-rank matrix completion such as one-bit matrix completion, it is still unclear whether the global optimality holds for all local minima.

## 3  Low-Rank Matrix Estimation

In this section, we provide a general problem setup for low-rank matrix estimation, together with several illustrative examples to show the applicability of our general framework.

### 3.1  General Problem Setup

Let $\mathbf{X}^* \in \mathbb{R}^{d_1 \times d_2}$ be an unknown low-rank matrix with rank $r$. Our goal is to estimate $\mathbf{X}^*$ through a collection of $n$ observations. Let $\mathcal{L}_n : \mathbb{R}^{d_1 \times d_2} \to \mathbb{R}$ be the sample loss function, which measures the fitness of any matrix $\mathbf{X}$ with respect to the given observations. Thus, the low-rank matrix estimation can be formulated as the following optimization problem

$$\min_{\mathbf{X} \in \mathbb{R}^{d_1 \times d_2}} \mathcal{L}_n(\mathbf{X}), \text{ subject to } \mathbf{X} \in \mathcal{C}, \text{ rank}(\mathbf{X}) \leq r,$$

where $\mathcal{C} \subseteq \mathbb{R}^{d_1 \times d_2}$ is a feasible set, such that $\mathbf{X}^* \in \mathcal{C}$. Note that when the observations are noisy, $\mathbf{X}^*$ is no longer the minimizer of $\mathcal{L}_n$. Hence, we further define an expected loss function $\bar{\mathcal{L}}_n(\mathbf{X})$ as the expectation of $\mathcal{L}_n(\mathbf{X})$ with respect to the randomness of our model (e.g., noise), i.e., $\bar{\mathcal{L}}_n(\mathbf{X}) = \mathbb{E}[\mathcal{L}_n(\mathbf{X})]$. In this way, the unknown low-rank matrix $\mathbf{X}^*$ is indeed the minimizer of the expected loss function $\bar{\mathcal{L}}_n$.

In order to solve the low-rank matrix estimation problem more efficiently, following Jain et al. (2013); Tu et al. (2015); Zheng and Lafferty (2016); Park et al. (2016a), we reparameterize $\mathbf{X}$ as $\mathbf{U}\mathbf{V}^\top$, and solve the following nonconvex optimization problem

$$\min_{\substack{\mathbf{U} \in \mathbb{R}^{d_1 \times r} \\ \mathbf{V} \in \mathbb{R}^{d_2 \times r}}} \mathcal{L}_n(\mathbf{U}\mathbf{V}^\top), \text{ subject to } \mathbf{U} \in \mathcal{C}_1, \mathbf{V} \in \mathcal{C}_2, \tag{3.1}$$

where $\mathcal{C}_1 \subseteq \mathbb{R}^{d_1 \times r}, \mathcal{C}_2 \subseteq \mathbb{R}^{d_2 \times r}$ are the corresponding rotation-invariant[1] feasible sets implied by $\mathcal{C}$. Suppose $\mathbf{X}^*$ can be decomposed as $\mathbf{X}^* = \mathbf{U}^* \mathbf{V}^{*\top}$, we need to ensure that $\mathbf{U}^* \in \mathcal{C}_1$ and $\mathbf{V}^* \in \mathcal{C}_2$.

## 3.2 Illustrative Examples

Here we briefly introduce matrix regression, matrix completion and one-bit matrix completion as three examples, to demonstrate the applicability of our generic framework.

### 3.2.1 Matrix Regression

In matrix regression (Recht et al., 2010; Negahban and Wainwright, 2011), our goal is to estimate the unknown rank-$r$ matrix $\mathbf{X}^* \in \mathbb{R}^{d_1 \times d_2}$ based on a set of noisy measurements $\boldsymbol{y} = \mathcal{A}(\mathbf{X}^*) + \boldsymbol{\epsilon}$, where $\mathcal{A} : \mathbb{R}^{d_1 \times d_2} \to \mathbb{R}^n$ is a linear operator such that $\mathcal{A}(\mathbf{X}^*) = (\langle \mathbf{A}_1, \mathbf{X}^* \rangle, \langle \mathbf{A}_2, \mathbf{X}^* \rangle, \ldots, \langle \mathbf{A}_n, \mathbf{X}^* \rangle)^\top$, and $\boldsymbol{\epsilon}$ is a.noise vector with i.i.d. sub-Gaussian entries with parameter $\nu$. Specifically, each random matrix $\mathbf{A}_i \in \mathbb{R}^{d_1 \times d_2}$ has i.i.d. standard normal entries. As discussed before, in order to estimate the low-rank matrix more efficiently, we consider the following nonconvex optimization problem

$$\min_{\substack{\mathbf{U} \in \mathbb{R}^{d_1 \times r} \\ \mathbf{V} \in \mathbb{R}^{d_2 \times r}}} \mathcal{L}_n(\mathbf{U}\mathbf{V}^\top) := \frac{1}{2n} \|\mathbf{y} - \mathcal{A}(\mathbf{U}\mathbf{V}^\top)\|_2^2.$$

Note that here the convex feasible sets $\mathcal{C}_1$ and $\mathcal{C}_2$ in (3.1) are both $\mathbb{R}^{d_1 \times r}$, which give rise to an unconstrained optimization.

### 3.2.2 Matrix Completion

In the noisy matrix completion (Rohde et al., 2011; Koltchinskii et al., 2011; Negahban and Wainwright, 2012), our goal is to recover the unknown rank-$r$ matrix $\mathbf{X}^* \in \mathbb{R}^{d_1 \times d_2}$ based on a set of randomly observed noisy entries from $\mathbf{X}^*$. For instance, one uniformly observes each entry independently with probability $p \in (0, 1)$. Specifically, we represent these observations by a random matrix $\mathbf{Y} \in \mathbb{R}^{d_1 \times d_2}$ such that

$$Y_{jk} := \begin{cases} X_{jk}^* + Z_{jk}, & \text{with probability } p, \\ *, & \text{otherwise,} \end{cases}$$

---

[1] We say $\mathcal{C}_1$ is rotation-invariant, if for any $\mathbf{A} \in \mathcal{C}_1$, $\mathbf{A}\mathbf{R} \in \mathcal{C}_1$, where $\mathbf{R}$ is an arbitrary $r$-by-$r$ orthogonal matrix.

where $\mathbf{Z} = (Z_{jk}) \in \mathbb{R}^{d_1 \times d_2}$ is a noise matrix with i.i.d. entries, such that each entry $Z_{jk}$ follows sub-Gaussian distribution with parameter $\nu$. Let $\Omega \subseteq [d_1] \times [d_2]$ be the index set of the observed entries, then we can estimate the low-rank matrix $\mathbf{X}^*$ by solving the following nonconvex optimization problem

$$\min_{\substack{\mathbf{U} \in \mathbb{R}^{d_1 \times r} \\ \mathbf{V} \in \mathbb{R}^{d_2 \times r}}} \mathcal{L}_\Omega(\mathbf{U}\mathbf{V}^\top) := \frac{1}{2p} \sum_{(j,k) \in \Omega} (\mathbf{U}_{j*}\mathbf{V}_{k*}^\top - Y_{jk})^2,$$

where $p = |\Omega|/(d_1 d_2)$. Here the feasible sets $\mathcal{C}_1$ and $\mathcal{C}_2$ in (3.1) are defined as follow

$$\mathcal{C}_i = \big\{ \mathbf{A} \in \mathbb{R}^{d_i \times r} \, \big| \, \|\mathbf{A}\|_{2,\infty} \leq \gamma \big\},$$

where $i \in \{1, 2\}$, and $\gamma > 0$ is a constant, which will be defined in later analysis.

### 3.2.3 One-Bit Matrix Completion

In one-bit matrix completion (Davenport et al., 2014; Cai and Zhou, 2013), we observe the sign of a random subset of noisy entries from the unknown rank-$r$ matrix $\mathbf{X}^* \in \mathbb{R}^{d_1 \times d_2}$, instead of observing the actual entries. In particular, we consider one-bit matrix completion problem under the uniform random sampling model (Davenport et al., 2014; Cai and Zhou, 2013; Ni and Gu, 2016). Given a probability density function $f : \mathbb{R} \to [0, 1]$ and an index set $\Omega \subseteq [d_1] \times [d_2]$, we observe the corresponding set of entries from a binary matrix $\mathbf{Y}$ according to the following probabilistic model:

$$Y_{jk} = \begin{cases} +1, & \text{with probability } f(X_{jk}^*), \\ -1, & \text{with probability } 1 - f(X_{jk}^*). \end{cases} \tag{3.2}$$

If $f$ is the cumulative distribution function of $-Z_{jk}$, where $\mathbf{Z} = (Z_{jk}) \in \mathbb{R}^{d_1 \times d_2}$ is a noise matrix with i.i.d. entries, then we can rewrite the above model as

$$Y_{jk} = \begin{cases} +1, & \text{if } X_{jk}^* + Z_{jk} > 0, \\ -1, & \text{if } X_{jk}^* + Z_{jk} < 0. \end{cases} \tag{3.3}$$

One widely-used probability density function is the logistic function $f(X_{jk}) = e^{X_{jk}}/(1 + e^{X_{jk}})$, which is equivalent to the fact that $Z_{jk}$ in (3.3) follows the standard logistic distribution. Given the probability density function $f$, the negative log-likelihood is given by

$$\mathcal{L}_\Omega(\mathbf{X}) := -\frac{1}{p} \sum_{(j,k) \in \Omega} \left\{ \mathbb{1}_{(Y_{jk}=1)} \log \big( f(X_{jk}) \big) + \mathbb{1}_{(Y_{jk}=-1)} \log \big( 1 - f(X_{jk}) \big) \right\},$$

where $p = |\Omega|/(d_1 d_2)$. Similar to the previous case, we can efficiently estimate $\mathbf{X}^*$ by solving a nonconvex optimization problem through matrix factorization.

## 4 The Proposed Algorithm

In this section, we propose an optimization algorithm to solve (3.1) based on gradient descent. It is important to note that the optimal solution to (3.1) is not unique. To be specific, for any solution $(\mathbf{U}, \mathbf{V})$ to the optimization problem (3.1), $\big( \mathbf{UP}, \mathbf{V}(\mathbf{P}^{-1})^\top \big)$ is also a valid solution, where $\mathbf{P} \in \mathbb{R}^{r \times r}$ can be any invertible matrix. In order to address this issue, following Tu et al. (2015); Zheng and Lafferty (2016); Park et al. (2016b), we consider the following optimization problem, which has an additional regularizer to force the two factors to be "balanced":

$$\min_{\substack{\mathbf{U} \in \mathbb{R}^{d_1 \times r} \\ \mathbf{V} \in \mathbb{R}^{d_2 \times r}}} \mathcal{L}_n(\mathbf{U}\mathbf{V}^\top) + \frac{1}{8}\|\mathbf{U}^\top\mathbf{U} - \mathbf{V}^\top\mathbf{V}\|_F^2, \quad \text{subject to} \quad \mathbf{U} \in \mathcal{C}_1, \mathbf{V} \in \mathcal{C}_2. \tag{4.1}$$

We present a gradient descent algorithm to solve the proposed estimator in (4.1), which is displayed in Algorithm 1.

---

**Algorithm 1** Gradient Descent (GD)

---

1: **Input:** Loss function $\mathcal{L}_n$, step size $\eta$, number of iterations $T$, initial solutions $\mathbf{U}^0, \mathbf{V}^0$.
2:     **for:** $t = 0, 1, 2, \ldots, T - 1$ **do**
3:         $\mathbf{U}^{t+1} = \mathbf{U}^t - \eta\big(\nabla_U \mathcal{L}_n(\mathbf{U}^t\mathbf{V}^{t\top}) - \frac{1}{2}\mathbf{U}^t(\mathbf{U}^{t\top}\mathbf{U}^t - \mathbf{V}^{t\top}\mathbf{V}^t)\big)$
4:         $\mathbf{V}^{t+1} = \mathbf{V}^t - \eta\big(\nabla_V \mathcal{L}_n(\mathbf{U}^t\mathbf{V}^{t\top}) - \frac{1}{2}\mathbf{V}^t(\mathbf{V}^{t\top}\mathbf{V}^t - \mathbf{U}^{t\top}\mathbf{U}^t)\big)$
5:         $\mathbf{U}^{t+1} = \mathcal{P}_{\mathcal{C}_1}(\mathbf{U}^{t+1})$
6:         $\mathbf{V}^{t+1} = \mathcal{P}_{\mathcal{C}_2}(\mathbf{V}^{t+1})$
7:     **end for**
8: **Output:** $\mathbf{X}^T = \mathbf{U}^T\mathbf{V}^{T\top}$

---

Here $\mathcal{P}_{\mathcal{C}_i}$ denotes the projection operator onto the feasible set $\mathcal{C}_i$, where $i \in \{1, 2\}$. Algorithm 1 is more general than Tu et al. (2015); Zheng and Lafferty (2015, 2016), because it applies to a larger family of loss functions. Therefore, various low-rank matrix estimation problems including those examples discussed in Section 3.2 can be solved by Algorithm 1. Compared with the algorithm proposed by Park et al. (2016b), we include a projection step to ensure the estimators lie in a feasible set, which is essential for many low-rank matrix recovery problems such as matrix completion and one-bit matrix completion.

As will be seen in our theoretical analysis, it is guaranteed to converge to the true parameters $\mathbf{U}^*$ and $\mathbf{V}^*$, only if the initial solutions $\mathbf{U}^0$ and $\mathbf{V}^0$ are sufficiently close to $\mathbf{U}^*$ and $\mathbf{V}^*$. Thus, inspired by Jain et al. (2010), we propose an initialization algorithm, which is displayed in Algorithm 2, to satisfy this requirement. For any matrix $\mathbf{X} \in \mathbb{R}^{d_1 \times d_2}$, we denote its rank-$r$ singular value decomposition by $\text{SVD}_r(\mathbf{X})$. Moreover, if $\text{SVD}_r(\mathbf{X}) = [\mathbf{U}, \mathbf{\Sigma}, \mathbf{V}]$, then we denote the best rank-$r$ approximation of $\mathbf{X}$ by $\mathcal{P}_r(\mathbf{X}) = \mathbf{U}\mathbf{\Sigma}\mathbf{V}^\top$, where $\mathcal{P}_r$ is a projection operator onto the rank-$r$ subspace.

---

**Algorithm 2** Initialization

---

1: **Input:** Loss function $\mathcal{L}_n$, parameter $\tau$, number of iterations S.
2:     $\mathbf{X}_0 = 0$
3:     **for:** $s = 1, 2, 3, \ldots, S$ **do**
4:         $\mathbf{X}_s = \mathcal{P}_r\big(\mathbf{X}_{s-1} - \tau\nabla\mathcal{L}_n(\mathbf{X}_{s-1})\big)$
5:     **end for**
6:     $[\overline{\mathbf{U}}^0, \mathbf{\Sigma}^0, \overline{\mathbf{V}}^0] = \text{SVD}_r(\mathbf{X}_S)$
7:     $\mathbf{U}^0 = \overline{\mathbf{U}}^0(\mathbf{\Sigma}^0)^{1/2}, \mathbf{V}^0 = \overline{\mathbf{V}}^0(\mathbf{\Sigma}^0)^{1/2}$
8: **Output:** $\mathbf{U}^0, \mathbf{V}^0$

---

Combining Algorithms 1 and 2, it is guaranteed that our gradient descent algorithm achieves linear convergence rate.

# 5 Main Theory

In this section, we are going to present our main theoretical results for the proposed algorithms. To begin with, we introduce some notations and facts to simplify our proof.

Let the singular value decomposition (SVD) of $\mathbf{X}^*$ be $\mathbf{X}^* = \overline{\mathbf{U}}^* \mathbf{\Sigma}^* \overline{\mathbf{V}}^{*\top}$, where $\overline{\mathbf{U}}^* \in \mathbb{R}^{d_1 \times r}$, $\overline{\mathbf{V}}^* \in \mathbb{R}^{d_2 \times r}$ are orthonormal matrices, and $\mathbf{\Sigma}^* \in \mathbb{R}^{r \times r}$ is a diagonal matrix. Let $\sigma_1 \geq \sigma_2 \geq \cdots \geq \sigma_r \geq 0$ be the sorted nonzero singular values of $\mathbf{X}^*$, and denote the condition number of $\mathbf{X}^*$ by $\kappa$, i.e., $\kappa = \sigma_1 / \sigma_r$. Besides, let $\mathbf{U}^* = \overline{\mathbf{U}}^* (\mathbf{\Sigma}^*)^{1/2}$ and $\mathbf{V}^* = \overline{\mathbf{V}}^* (\mathbf{\Sigma}^*)^{1/2}$, then following [Tu et al. (2015)](#); [Zheng and Lafferty (2016)](#), we can lift the low-rank matrix $\mathbf{X}^* \in \mathbb{R}^{d_1 \times d_2}$ to a positive semidefinite matrix $\mathbf{Y}^* \in \mathbb{R}^{(d_1+d_2) \times (d_1+d_2)}$ in higher dimension

$$\mathbf{Y}^* = \begin{bmatrix} \mathbf{U}^* \mathbf{U}^{*\top} & \mathbf{U}^* \mathbf{V}^{*\top} \\ \mathbf{V}^* \mathbf{U}^{*\top} & \mathbf{V}^* \mathbf{V}^{*\top} \end{bmatrix} = \mathbf{Z}^* \mathbf{Z}^{*\top},$$

where $\mathbf{Z}^*$ is defined as

$$\mathbf{Z}^* = \begin{bmatrix} \mathbf{U}^* \\ \mathbf{V}^* \end{bmatrix} \in \mathbb{R}^{(d_1+d_2) \times r}.$$

Observant readers may have already noticed that the symmetric factorization of $\mathbf{Y}^*$ is not unique. In order to address this issue, it is convenient to define a solution set, which can be seen as an equivalent class of the optimal solutions. Thus, we define the solution sets with respect to the true parameter $\mathbf{Z}^*$ as

$$\mathcal{Z} = \left\{ \mathbf{Z} \in \mathbb{R}^{(d_1+d_2) \times r} \,\middle|\, \mathbf{Z} = \mathbf{Z}^* \mathbf{R} \text{ for some } \mathbf{R} \in \mathbb{Q}_r \right\},$$

where $\mathbb{Q}_r$ denotes the set of $r$-by-$r$ orthonormal matrices. Note that for any $\mathbf{Z} \in \mathcal{Z}$, we have $\mathbf{X}^* = \mathbf{Z}_U \mathbf{Z}_V^\top$, where $\mathbf{Z}_U$ and $\mathbf{Z}_V$ denote the top $d_1$ and bottom $d_2$ rows of matrix $\mathbf{Z} \in \mathbb{R}^{(d_1+d_2) \times r}$, respectively.

**Definition 5.1.** Define the estimation error $d(\mathbf{Z}, \mathbf{Z}^*)$ as the minimal Frobenius norm between $\mathbf{Z}$ and $\mathbf{Z}^*$ with respect to the optimal rotation, namely

$$d(\mathbf{Z}, \mathbf{Z}^*) = \min_{\widetilde{\mathbf{Z}} \in \mathcal{Z}} \|\mathbf{Z} - \widetilde{\mathbf{Z}}\|_F = \min_{\mathbf{R} \in \mathbb{Q}_r} \|\mathbf{Z} - \mathbf{Z}^* \mathbf{R}\|_F.$$

**Definition 5.2.** We denote the local region around optimum $\mathbf{Z}^*$ with radius $R$ as

$$\mathbb{B}(R) = \left\{ \mathbf{Z} \in \mathbb{R}^{(d_1+d_2) \times r} \,\middle|\, d(\mathbf{Z}, \mathbf{Z}^*) \leq R \right\}.$$

Before we present our main results, we first lay out several necessary conditions regarding $\mathcal{L}_n$ ane $\bar{\mathcal{L}}_n$. First, we impose two conditions on the sample loss function $\mathcal{L}_n$. These two conditions are known as restricted strongly convexity (RSC) and restricted strongly smoothness (RSM) conditions ([Negahban et al., 2009](#); [Loh and Wainwright, 2013](#)), assuming that there exist both quadratic lower bound and upper bound, respectively, on the remaining term of the first order Taylor expansion of $\mathcal{L}_n$.

**Condition 5.3** (Restricted Strongly Convexity). For a given sample size $n$, $\mathcal{L}_n$ is restricted strongly convex with parameter $\mu$, such that for any rank-$r$ matrices $\mathbf{X}, \mathbf{Y} \in \mathbb{R}^{d_1 \times d_2}$

$$\mathcal{L}_n(\mathbf{Y}) \geq \mathcal{L}_n(\mathbf{X}) + \langle \nabla \mathcal{L}_n(\mathbf{X}), \mathbf{Y} - \mathbf{X} \rangle + \frac{\mu}{2} \|\mathbf{Y} - \mathbf{X}\|_F^2.$$

**Condition 5.4** (Restricted Strongly Smoothness). Given a fixed sample size $n$, $\mathcal{L}_n$ is restricted strongly smooth with parameter $L$, such that for any rank-$r$ matrices $\mathbf{X}, \mathbf{Y} \in \mathbb{R}^{d_1 \times d_2}$

$$\mathcal{L}_n(\mathbf{Y}) \leq \mathcal{L}_n(\mathbf{X}) + \langle \nabla \mathcal{L}_n(\mathbf{X}), \mathbf{Y} - \mathbf{X} \rangle + \frac{L}{2} \|\mathbf{Y} - \mathbf{X}\|_F^2.$$

Both Conditions 5.3 and 5.4 can be verified for each of our illustrative examples, discussed in Section 3.2.

Moreover, we impose similar strongly convexity and smoothness conditions on the expected loss function $\bar{\mathcal{L}}_n$, which is the expectation of the sample loss function with respect to the randomness of our model (e.g., noise).

**Condition 5.5** (Restricted Strongly Convexity). For a given sample size $n$, $\bar{\mathcal{L}}_n$ is restricted strongly convex with parameter $\bar{\mu}$, such that for any rank-$r$ matrices $\mathbf{X}, \mathbf{Y} \in \mathbb{R}^{d_1 \times d_2}$

$$\bar{\mathcal{L}}_n(\mathbf{Y}) \geq \bar{\mathcal{L}}_n(\mathbf{X}) + \langle \nabla \bar{\mathcal{L}}_n(\mathbf{X}), \mathbf{Y} - \mathbf{X} \rangle + \frac{\bar{\mu}}{2} \|\mathbf{Y} - \mathbf{X}\|_F^2.$$

**Condition 5.6** (Restricted Strongly Smoothness). Given a fixed sample size $n$, $\bar{\mathcal{L}}_n$ is restricted strongly smooth with parameter $\bar{L}$, such that for any rank-$r$ matrices $\mathbf{X}, \mathbf{Y} \in \mathbb{R}^{d_1 \times d_2}$

$$\bar{\mathcal{L}}_n(\mathbf{Y}) \leq \bar{\mathcal{L}}_n(\mathbf{X}) + \langle \nabla \bar{\mathcal{L}}_n(\mathbf{X}), \mathbf{Y} - \mathbf{X} \rangle + \frac{\bar{L}}{2} \|\mathbf{Y} - \mathbf{X}\|_F^2.$$

We will further demonstrate in our illustrative examples that the expected loss function $\bar{\mathcal{L}}_n$ indeed satisfies Conditions 5.5 and 5.6.

Finally, we assume that the difference between the gradient of the sample loss function $\nabla \mathcal{L}_n$ and that of the expected loss function $\nabla \bar{\mathcal{L}}_n$ is upper bounded in terms of spectral norm.

**Condition 5.7.** For a given sample size $n$ and tolerance parameter $\delta \in (0, 1)$, we let $\epsilon(n, \delta)$ be the smallest scalar such that, for any fixed $\mathbf{X} \in \mathbb{R}^{d_1 \times d_2}$, with probability at least $1 - \delta$, we have

$$\|\nabla \mathcal{L}_n(\mathbf{X}) - \nabla \bar{\mathcal{L}}_n(\mathbf{X})\|_2 \leq \epsilon(n, \delta),$$

where $\epsilon(n, \delta)$ depends on sample size $n$ and $\delta$.

Condition 5.7 is essential to derive the statistical error bound regarding the estimator returned by our algorithm.

## 5.1 Results for the Generic Model

Here, we first provide theoretical guarantees of our proposed algorithm for the generic model, where the loss function is any loss function $\mathcal{L}(\mathbf{X})$ that satisfies the above conditions.

**Theorem 5.8** (Gradient Descent). Recall that $\mathbf{X}^* = \mathbf{U}^* \mathbf{V}^{*\top}$ is an unknown rank-$r$ matrix. For any $\mathbf{Z}^0 \in \mathbb{B}(c_2\sqrt{\sigma_r})$, where $c_2 \leq \min\{1/4, \sqrt{2\bar{\mu}'/\{5(4\bar{L}+1)\}}\}$, if the sample size $n$ is large enough such that

$$\epsilon^2(n, \delta) \leq \frac{c_2^2 \bar{\mu}' \sigma_r^2}{10 c_3 r},$$

where $\bar{\mu}' = \min\{\bar{\mu}, 1\}$ and $c_3 = 2/\bar{L} + 4/\bar{\mu}$, then with step size $\eta = c_1/\sigma_1$, where $c_1 \leq \min\{1/(64\bar{L}), 1/32\}$, the estimator at iteration $t$ of Algorithm 1 satisfies

$$d^2(\mathbf{Z}^{t+1}, \mathbf{Z}^*) \leq \left(1 - \frac{c_1\bar{\mu}'}{10\kappa}\right)d^2(\mathbf{Z}^t, \mathbf{Z}^*) + \eta c_3 r \epsilon^2(n, \delta),$$

with probability at least $1 - \delta$. If we let $\rho = 1 - c_1\bar{\mu}'/(10\kappa)$, then the iterates $\{\mathbf{Z}^t\}_{t=0}^{\infty}$ satisfy

$$d^2(\mathbf{Z}^t, \mathbf{Z}^*) \leq \rho^t d^2(\mathbf{Z}^0, \mathbf{Z}^*) + \frac{10c_3 r}{\bar{\mu}'\sigma_r}\epsilon^2(n, \delta),$$

with probability at least $1 - \delta$.

Thus, it is sufficient to perform $T = O\big(\kappa \log(1/\epsilon)\big)$ iterations for $\mathbf{Z}^T$ to converge to a close neighborhood of $\mathbf{Z}^*$, where $\epsilon$ depends on the statistical error term $r\epsilon^2(n, \delta)$. Note that in Theorem 5.8, the step size $\eta$ is chosen according to $1/\sigma_1$. In practice, we can set the step size as $\eta = c'/\|\mathbf{Z}^0\|_2^2$, where $c'$ is a small constant, since $\sqrt{\sigma_1} \leq \|\mathbf{Z}^0\|_2 \leq 2\sqrt{\sigma_1}$ holds as long as $\mathbf{Z}^0 \in \mathbb{B}(\sqrt{\sigma_r}/4)$. Moreover, the reconstruction error $\|\mathbf{X}^T - \mathbf{X}^*\|_F^2$ can be upper bounded by $C\sigma_1 d^2(\mathbf{Z}^T, \mathbf{Z}^*)$, where $C$ is a universal constant. Therefore, $\mathbf{X}^T$ is indeed a good estimator for $\mathbf{X}^*$.

**Theorem 5.9** (Initialization). Recall that $\mathbf{X}^* = \mathbf{U}^*\mathbf{V}^{*\top}$ is an unknown rank-$r$ matrix. Consider $\mathbf{U}^0$, $\mathbf{V}^0$ produced in the initialization Algorithm 2, and let $\mathbf{X}^0 = \mathbf{U}^0\mathbf{V}^{0\top}$. If $L/\mu \in (1, 4/3)$, then with step size $\tau = 1/L$, we have

$$\|\mathbf{X}^0 - \mathbf{X}^*\|_F \leq \rho^S\|\mathbf{X}^*\|_F + \frac{2\sqrt{3r}\epsilon(n, \delta)}{L(1 - \rho)},$$

with probability at least $1 - \delta$, where $\rho = 2\sqrt{1 - \mu/L}$ is the contraction parameter.

**Remark 5.10.** In order to satisfy the initial assumption $\mathbf{Z}^0 \in \mathbb{B}(c_2\sqrt{\sigma_r})$ in Theorem 5.8, it is sufficient to make sure $\mathbf{X}^0$ is close enough to the unknown rank-$r$ matrix $\mathbf{X}^*$, i.e., $\|\mathbf{X}^0 - \mathbf{X}^*\|_F \leq c\sigma_r$, where $c \leq \min\{1/2, 2c_2\}$. In fact, since $\|\mathbf{X}^0 - \mathbf{X}^*\|_2 \leq \|\mathbf{X}^0 - \mathbf{X}^*\|_F \leq \sigma_r/2$, we have

$$d^2(\mathbf{Z}, \mathbf{Z}^*) \leq \frac{\sqrt{2} - 1}{2}\frac{\|\mathbf{X}^0 - \mathbf{X}^*\|_F^2}{\sigma_r} \leq c_2^2\sigma_r.$$

Therefore, we need to assume the sample size $n$ to be large enough such that

$$\epsilon(n, \delta) \leq \frac{cL(1 - \rho)\sigma_r}{2\sqrt{3r}},$$

which has the same order as the error bound in Theorem 5.8, and it is sufficient to perform $S = O(1)$ iterations in Algorithm 2 to make sure $\|\mathbf{X}^0 - \mathbf{X}^*\|_F \leq c\sigma_r$. Furthermore, our initialization algorithm only requires the condition $L/\mu \in (1, 4/3)$, which significantly relaxes the strict condition required in Park et al. (2016b), i.e.,

$$\frac{L}{\mu} \leq 1 + \frac{\sigma_r^2}{4608\|\mathbf{X}^*\|_F^2}.$$

## 5.2 Results for Specific Examples

The deterministic results in Theorem 5.8 are fairly abstract in nature. Here, we consider the specific examples of low-rank matrix estimation in Section 3.2, and demonstrate how to apply our general results in Section 5.1 to these examples. In the following discussions, we denote $d' = \max\{d_1, d_2\}$.

### 5.2.1 Matrix Regression

We analyze the example of matrix regression. First, we obtain the restricted strongly convexity and smoothness parameters $\mu = 4/9$ and $L = 5/9$ for both $\mathcal{L}_n$ and $\bar{\mathcal{L}}_n$. Moreover, we derive the upper bound of the difference between $\nabla \mathcal{L}_n$ and $\nabla \bar{\mathcal{L}}_n$. Finally, we provide theoretical guarantees for our algorithm under the matrix regression model.

**Corollary 5.11.** Under the previously stated conditions, consider the estimator $\mathbf{Z}^t$ produced at iteration $t$ in Algorithm 1. Then there exist constants $c_1, c_2, c_3, c_4$ and $c_5$, such that with step size $\eta \leq c_1/\sigma_1$ and initial solution $\mathbf{Z}^0$ satisfying $\mathbf{Z}^0 \in \mathbb{B}(c_2\sqrt{\sigma_r})$, we have

$$d^2(\mathbf{Z}^{t+1}, \mathbf{Z}^*) \leq \left(1 - \frac{2\sigma_r \eta}{45}\right) d^2(\mathbf{Z}^t, \mathbf{Z}^*) + \eta c_5 \nu^2 \frac{rd'}{n},$$

holds with probability at least $1 - c_3 \exp\big(-c_4 d'\big)$.

**Remark 5.12.** In the noisy case, Corollary 5.11 suggests that, after $O(\kappa \log(n/(rd')))$ number of iterations, the output of our algorithm achieves $O(\sqrt{rd'/n})$ statistical error, which matches the minimax lower bound for matrix sensing (Negahban and Wainwright, 2011). While in the noiseless case, in order to satisfy restricted strongly convexity and smoothness conditions, we require the sample size $n = O(rd')$, which achieves the optimal sample complexity for matrix sensing (Recht et al., 2010; Tu et al., 2015).

### 5.2.2 Matrix Completion

In matrix completion, we consider a partially observed setting, such that we only observe entries of $\mathbf{X}^*$ over a subset $\mathcal{X} \subseteq [d_1] \times [d_2]$. We assume a uniform sampling model such that

$$\forall (j, k) \in \mathcal{X}, \ j \sim \text{uniform}([d_1]), \ k \sim \text{uniform}([d_2]).$$

It is observed in Gross (2011) that if $\mathbf{X}^*$ is equal to zero in nearly all elements, it is impossible to recovery $\mathbf{X}^*$ unless all of its entries are sampled. In other words, there will always be some low-rank matrices, which are too spiky (Negahban and Wainwright, 2012; Gunasekar et al., 2014) to be recovered without sampling the whole matrix. In order to avoid the overly spiky matrices in matrix completion, we add an infinity norm constraint $\|\mathbf{X}^*\|_\infty \leq \alpha$ into our estimator, which is known as spikiness condition (Negahban and Wainwright, 2012). It is argued that the spikiness condition is much less restricted than the incoherence conditions (Candès and Recht, 2009) imposed in exact low-rank matrix completion (Negahban and Wainwright, 2012; Klopp et al., 2014).

Therefore, we consider the class of low-rank matrices with infinity norm constraint as follows $\mathcal{C}(\alpha) = \big\{\mathbf{X} \in \mathbb{R}^{d_1 \times d_2} \mid \|\mathbf{X}\|_\infty \leq \alpha\big\}$. Based on $\mathcal{C}(\alpha)$, we further define feasible sets $\mathcal{C}_i = \{\mathbf{A} \in \mathbb{R}^{d_i \times r} \mid \|\mathbf{A}\|_{2,\infty} \leq \sqrt{\alpha}\}$, where $i \in \{1, 2\}$. In this way, for any $\mathbf{U} \in \mathcal{C}_1$ and $\mathbf{V} \in \mathcal{C}_2$, we have $\mathbf{U}\mathbf{V}^\top \in \mathcal{C}(\alpha)$. By imposing spikiness condition, we can establish the restricted strongly convexity and smoothness condition for $\mathcal{L}_n$ and $\bar{\mathcal{L}}_n$ with parameters $\mu = 8/9$ and $L = 10/9$. Moreover, we obtain the upper bound of the difference between $\nabla \mathcal{L}_n$ and $\nabla \bar{\mathcal{L}}_n$. Finally, we provide theoretical guarantees for our algorithm under the matrix completion model.

**Corollary 5.13.** Under the previously stated conditions, suppose $\mathbf{X}^* \in \mathcal{C}(\alpha)$, for the estimator $\mathbf{Z}^t$ produced at iteration $t$ in Algorithm 1, there exist constants $c_1, c_2, c_3$ and $c_4$, such that with step size $\eta \leq c_1/\sigma_1$ and initial solution $\mathbf{Z}^0$ satisfying $\mathbf{Z}^0 \in \mathbb{B}(c_2\sqrt{\sigma_r})$, we have

$$d^2(\mathbf{Z}^{t+1}, \mathbf{Z}^*) \leq \left(1 - \frac{4\sigma_r \eta}{45}\right) d^2(\mathbf{Z}^t, \mathbf{Z}^*) + \eta c_4 \max\{\nu^2, \alpha^2\} \frac{rd' \log d'}{p},$$

holds with probability at least $1 - c_3/d'$.

**Remark 5.14.** For matrix completion with noisy observations, Corollary 5.13 suggests that after $O(\kappa \log (n/(rd' \log d'))$ number of iterations, for the standardized error term $\|\mathbf{X}^T - \mathbf{X}^*\|_F/\sqrt{d_1 d_2}$, our algorithm attains $O(\sqrt{rd' \log d'/n})$ statistical error, which matches the minimax lower bound for matrix completion established in Negahban and Wainwright (2012); Koltchinskii et al. (2011). While in the noiseless case, in order to guarantee restricted strongly convexity and smoothness conditions, we require the sample size $n = O(rd' \log d')$, which obtains optimal sample complexity for matrix completion (Candès and Recht, 2009; Recht, 2011; Chen et al., 2013).

### 5.2.3 One-Bit Matrix Completion

For the example of one bit matrix completion, we establish the restricted strongly convexity and smoothness condition for both $\mathcal{L}_n$ and $\bar{\mathcal{L}}_n$. In this case, we have the strongly convexity and smoothness parameters $\mu = C_1 \mu_\alpha$ and $L = C_2 L_\alpha$, where $\mu_\alpha$ and $L_\alpha$ satisfy

$$\mu_\alpha \leq \min \left( \inf_{|x| \leq \alpha} \left\{ \frac{f'^2(x)}{f^2(x)} - \frac{f''(x)}{f(x)} \right\}, \inf_{|x| \leq \alpha} \left\{ \frac{f'^2(x)}{(1-f(x))^2} + \frac{f''(x)}{1-f(x)} \right\} \right), \tag{5.1}$$

$$L_\alpha \geq \max \left( \sup_{|x| \leq \alpha} \left\{ \frac{f'^2(x)}{f^2(x)} - \frac{f''(x)}{f(x)} \right\}, \sup_{|x| \leq \alpha} \left\{ \frac{f'^2(x)}{(1-f(x))^2} + \frac{f''(x)}{1-f(x)} \right\} \right), \tag{5.2}$$

where $\alpha$ is the upper bound of the absolute value for every entry $X_{jk}$, and $f(x)$ is the probability density function. When $\alpha$ is a fixed constant, and $\mathcal{L}_n(\cdot)$ is specified, $\mu_\alpha$ and $L_\alpha$ are fixed constants which do not depend on dimension. For instance, we have $\mu_\alpha = e^\alpha/(1 + e^\alpha)^2$ and $L_\alpha = 1/4$ for the logistic model. Another important quantity is $\gamma_\alpha$, which reflects the steepness of the function $\mathcal{L}_n(\cdot)$

$$\gamma_\alpha \geq \sup_{|x| \leq \alpha} \left\{ \frac{|f'(x)|}{f(x)\big(1-f(x)\big)} \right\}. \tag{5.3}$$

Furthermore, we obtain the upper bound of the difference between $\nabla \mathcal{L}_n$ and $\nabla \bar{\mathcal{L}}_n$. Finally, we provide theoretical guarantees under one-bit matrix completion model.

**Corollary 5.15.** Under the previously stated conditions, suppose $\mathbf{X}^* \in \mathcal{C}(\alpha)$, a subset $\Omega$ of entries of the unknown matrix $\mathbf{X}^*$ is uniformly sampled from the log-concave probability density function $f$, and the binary matrix $\mathbf{Y}$ in (3.2) is generated based on the probability density function $f$. For estimator $\mathbf{Z}^t$ produced in Algorithm 1, there exist constants $c_1, c_2, c_3$ and $c_4$ such that with step size $\eta \leq c_1/\sigma_1$, and initial solution $\mathbf{Z}^0$ established by Algorithm 2 satisfying $\mathbf{Z}^0 \in \mathbb{B}(c_2\sqrt{\sigma_r})$, with probability at least $1 - c_3/d'$.

$$d^2(\mathbf{Z}^{t+1}, \mathbf{Z}^*) \leq \left(1 - \frac{\mu\sigma_r\eta}{10}\right)d^2(\mathbf{Z}^t, \mathbf{Z}^*) + \eta c_4 \max\{\gamma_\alpha^2, \alpha^2\}\frac{rd'\log d'}{p}.$$

**Remark 5.16.** For one-bit matrix completion, Corollary 5.15 suggests that after $O(\kappa \log (n/(rd'\log d'))$ number of iterations, for the standardized error term $\|\mathbf{X}^T - \mathbf{X}^*\|_F/\sqrt{d_1 d_2}$, our algorithm obtains $O(\sqrt{rd'\log d'/n})$ statistical error, which matches the minimax lower bound of one-bit matrix completion problem provided by Davenport et al. (2014); Cai and Zhou (2013).

# 6 Numerical Experiments

In this section, we perform experiments on synthetic data to further illustrate the theoretical results of our method. We consider three approaches for initialization: (a) One step SVD of $\nabla \mathcal{L}_n(0)$ (One Step), which has been used in Bhojanapalli et al. (2015); Park et al. (2016a,b); (b) Random initialization (Random), which is suggested by Bhojanapalli et al. (2016); Park et al. (2016c); Ge et al. (2016); (c) Our proposed initialization Algorithm 2. We investigate the convergence rates of gradient descent under different initialization approaches, and evaluate the sample complexity that is required to recover the unknown low-rank matrices. All the results are based on 30 trails.

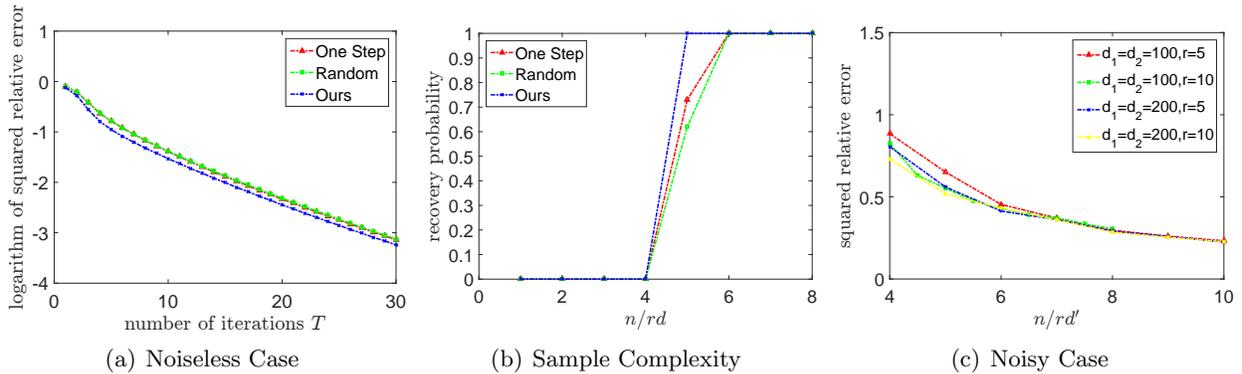

(a) Noiseless Case      (b) Sample Complexity      (c) Noisy Case

Figure 1: Simulation results for matrix regression. (a) Convergence rates for matrix regression in the noiseless case: logarithm of squared relative error $\|\widehat{\mathbf{X}} - \mathbf{X}^*\|_F^2/\|\mathbf{X}^*\|_F^2$ versus number of iterations, which implies the linear convergence rate of our algorithm; (b) Probability of successful recovery versus $n/(rd')$, which demonstrates the sample complexity scales linearly with $rd'$; (c) Statistical error for matrix regression in the noisy case: squared relative error $\|\widehat{\mathbf{X}} - \mathbf{X}^*\|_F^2/\|\mathbf{X}^*\|_F^2$ versus $n/(rd')$, which confirms the statistical error bound.

## 6.1 Matrix Regression

For matrix regression, we consider the unknown matrix $\mathbf{X}^*$ in the following settings: (i) $d_1 = 100, d_2 = 100, r = 5$; (ii) $d_1 = 100, d_2 = 100, r = 10$; (iii) $d_1 = 200, d_2 = 200, r = 5$; and (iv) $d_1 = 200, d_2 = 200, r = 10$. In all these settings, we first randomly generate $\mathbf{U}^* \in \mathbb{R}^{d_1 \times r}, \mathbf{V}^* \in$ $\mathbb{R}^{d_2 \times r}$ to obtain $\mathbf{X}^* = \mathbf{U}^* \mathbf{V}^{*\top}$. Next, we generate measurements based on the observation model $y_i = \langle \mathbf{A}_i, \mathbf{X}^* \rangle + \epsilon_i$, where each entry of observation matrix $\mathbf{A}_i$ follows i.i.d. standard Gaussian distribution. And we consider both (1) noisy case: the noise follows i.i.d. zero mean Gaussian distribution with standard deviation $\sigma = 0.1 \cdot \|\mathbf{X}^*\|_\infty$ and (2) noiseless case.

To illustrate the convergence rate, we report the squared relative error $\|\widehat{\mathbf{X}} - \mathbf{X}^*\|_F^2 / \|\mathbf{X}^*\|_F^2$ in terms of log scale. In the noiseless case, Figure 1(a) shows the linear convergence rates of our algorithm with different initializations under setting (i), which confirms the convergence results of our algorithm. To illustrate the sample complexity, we consider the empirical probability of successful recovery under different sample size. Based on the output $\widehat{\mathbf{X}}$ of our algorithm given $n$ random observations, a trial is considered to be successful, if the relative error is less than $10^{-3}$. The results of recovery probability under setting (i) with different initialization approaches are displayed in Figure 1(b). We conclude that there exists a phase transition around $n = 4rd'$, which confirms that the sample complexity $n$ is linear with the dimension $d'$ and the rank $r$. Besides, we obtain results with similar patterns for other settings, thus we leave them out for simplicity. Figure 1(c) demonstrates, in the noisy case with our proposed initialization, how the estimation errors scale with $n/(rd')$, which aligns well with our theory.

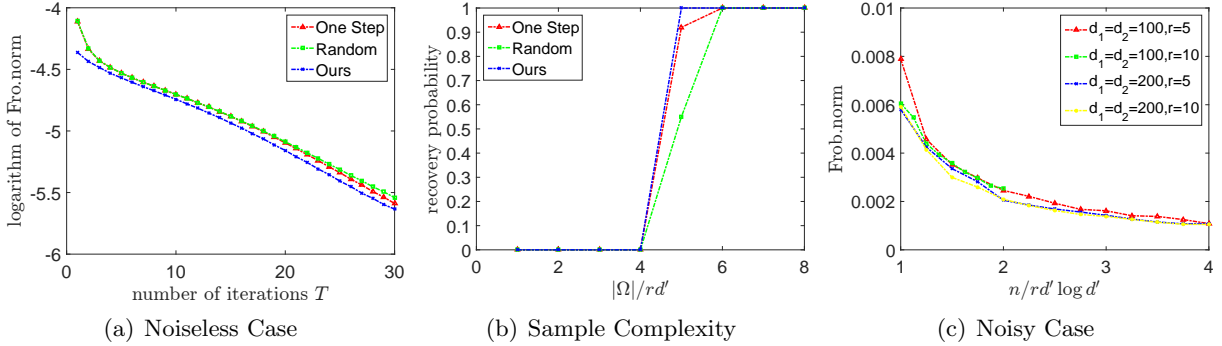

(a) Noiseless Case

(b) Sample Complexity

(c) Noisy Case

Figure 2: Simulation results for matrix completion. (a) Convergence rates for matrix completion in the noiseless case: logarithm of mean squared error in Frobenius norm $\|\widehat{\mathbf{X}} - \mathbf{X}^*\|_F^2 / (d_1 d_2)$ versus number of iterations, which implies the linear convergence rate of our algorithm; (b) Probability of successful recovery versus $n/(rd')$, which demonstrates the sample complexity scales linearly with $rd' \log d'$; (c) Statistical error for matrix completion in the noisy case: mean squared error in Frobenius norm $\|\widehat{\mathbf{X}} - \mathbf{X}^*\|_F^2 / (d_1 d_2)$ versus $n/(rd' \log d')$, which confirms the statistical error bound.

## 6.2 Matrix Completion

For matrix completion, the unknown matrix $\mathbf{X}^*$ is generated similarly as in matrix sensing, and the observation matrix $\mathbf{Y}$ are uniformly sampled. Besides, we consider both (1) noisy case: each entry of the noise matrix follows i.i.d. zero mean Gaussian distribution with standard deviation $\sigma = 0.1 \cdot \|\mathbf{X}^*\|_\infty$ and (2) noiseless case.

To illustrate the convergence rate, we compute the mean squared error in Frobenius norm $\|\widehat{\mathbf{X}} - \mathbf{X}^*\|_F^2 / (d_1 d_2)$. In the noiseless case, the linear convergence rates, under setting (i) with different initializations, are shown in Figure 2(a), which confirms our theoretical results. To

illustrate the sample complexity, we report the empirical recovery probability under setting (i) with different initializations in Figure 2(b). We conclude that there exists a phase transition around $n = 4rd'$, which confirms that the sample complexity $n$ is linear with the dimension $d'$ and the rank $r$. Besides, we obtain results with similar patterns for other settings, thus we leave them out for simplicity. In the noisy case, Figure 2(c) shows how the estimation errors scale with $n/(rd'\log d')$ with our proposed initialization, which is consistent with our theory.

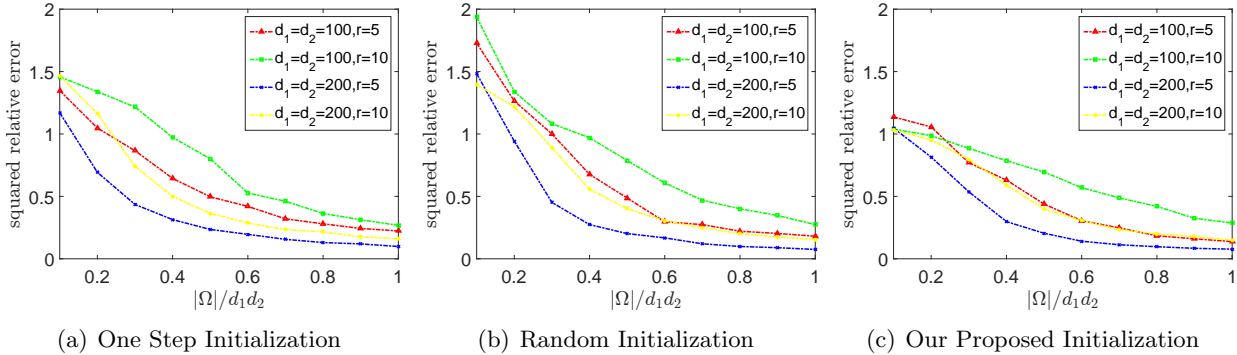

(a) One Step Initialization      (b) Random Initialization      (c) Our Proposed Initialization

Figure 3: Performance of proposed estimator for one bit matrix completion under different initialization approaches: squared relative error $\|\widehat{\mathbf{X}} - \mathbf{X}^*\|_F^2 / \|\mathbf{X}^*\|_F^2$ versus $|\Omega|/(d_1 d_2)$, which confirms the statistical rate. (a) One step initialization; (b) Random initialization; (c) Our proposed initialization

### 6.3 One-Bit Matrix Completion

For one-bit matrix completion, we consider the similar setting as in Davenport et al. (2014); Bhaskar and Javanmard (2015); Ni and Gu (2016). We first generate the unknown low-rank matrix $\mathbf{X}^* = \mathbf{U}^* \mathbf{V}^{*\top}$, where $\mathbf{U}^* \in \mathbb{R}^{d_1 \times r}, \mathbf{V}^* \in \mathbb{R}^{d_2 \times r}$ are randomly generated from a uniform distribution on $[-1/2, 1/2]$. Then we scale $\mathbf{X}^*$ to make $\|\mathbf{X}^*\|_\infty = \alpha = 1$. Here we consider the Probit model under uniform sampling, namely $f(X_{ij}) = \Phi(\mathbf{X}_{ij}/\sigma)$ in (3.2), where $\Phi$ is the CDF of the standard Gaussian distribution. We set dimension $d_1 = d_2 \in \{100, 200\}$, rank $r \in \{5, 10\}$, and noise $\sigma = 0.18$.

In order to measure the performance of our estimator, we use the squared relative error which is defined as $\|\widehat{\mathbf{X}} - \mathbf{X}^*\|_F^2 / \|\mathbf{X}^*\|_F^2$. The results are illustrated in Figure 3(a), 3(b), and 3(c). We infer that the squared relative error decreases as the percentage of observed entries increase in all settings. We also conclude that with the same percentage of observed entries, the squared relative error decreases as the dimension $d_1, d_2$ increases, which further confirms the statistical error rate.

## 7 Conclusions

In this paper, we developed a unified framework for estimating low-rank matrices, which integrates both optimization-theoretic and statistical analyses. Our algorithm and theory can be applied to low-rank matrix estimation based on both noisy observations and noiseless observations. In addition, we proposed a new initialization algorithm to provide a desired initial estimator, which outperforms existing initialization algorithms for nonconvex low-rank matrix estimation. Thorough experiments on synthetic data verified the advantages of our algorithm and theory.

# A    Proofs of the Main Theory

In this section, we present the proof of our main theoretical results. Before proceeding to the proof, we introduce the following notations. For any $\mathbf{Z} \in \mathbb{R}^{(d_1+d_2)\times r}$, we denote $\mathbf{Z} = [\mathbf{U}; \mathbf{V}]$, where $\mathbf{U} \in \mathbb{R}^{d_1 \times r}$ and $\mathbf{V} \in \mathbb{R}^{d_2 \times r}$. According to (4.1), our objective is to minimize the following loss function in terms of $\mathbf{Z}$

$$\widetilde{F}_n(\mathbf{Z}) = F_n(\mathbf{U}, \mathbf{V}) = \mathcal{L}_n(\mathbf{U}\mathbf{V}^\top) + \frac{1}{8}\|\mathbf{U}^\top\mathbf{U} - \mathbf{V}^\top\mathbf{V}\|_F^2.$$

Therefore, we have

$$\nabla \widetilde{F}_n(\mathbf{Z}) = \begin{bmatrix} \nabla_{\mathbf{U}}\mathcal{L}_n(\mathbf{U}\mathbf{V}^\top) + \frac{1}{2}\mathbf{U}(\mathbf{U}^\top\mathbf{U} - \mathbf{V}^\top\mathbf{V}) \\ \nabla_{\mathbf{V}}\mathcal{L}_n(\mathbf{U}\mathbf{V}^\top) + \frac{1}{2}\mathbf{V}(\mathbf{U}^\top\mathbf{U} - \mathbf{V}^\top\mathbf{V}) \end{bmatrix}. \tag{A.1}$$

## A.1    Proof of Theorem 5.8

In order to prove the convergence result, we need to make use of the following lemmas regarding $\widetilde{F}_n$, the sample loss function with regularization term. Lemmas A.1 and A.2 show that $\widetilde{F}_n$ satisfies the local curvature and smoothness conditions, respectively. We present the proofs in Sections C.1 and C.2, respectively.

**Lemma A.1** (Local Curvature Condition). Recall that $\mathbf{X}^* = \mathbf{U}^*\mathbf{V}^{*\top}$ is the unknown rank-$r$ matrix. Let $\mathbf{Z} \in \mathbb{R}^{(d_1+d_2)\times r}$ be any matrix with $\mathbf{Z} = [\mathbf{U}; \mathbf{V}]$, where $\mathbf{U} \in \mathbb{R}^{d_1 \times r}$ and $\mathbf{V} \in \mathbb{R}^{d_2 \times r}$. Let $\mathbf{R} = \operatorname{argmin}_{\widetilde{\mathbf{R}}\in\mathbb{Q}_r}\|\mathbf{Z} - \mathbf{Z}^*\widetilde{\mathbf{R}}\|_F$ be the optimal rotation with respect to $\mathbf{Z}$, and $\mathbf{H} = \mathbf{Z} - \mathbf{Z}^*\mathbf{R}$, then we have

$$\langle \nabla\widetilde{F}_n(\mathbf{Z}), \mathbf{H}\rangle \geq \frac{\bar{\mu}'\sigma_r}{10}\|\mathbf{H}\|_F^2 + \frac{1}{4\bar{L}}\|\nabla\bar{\mathcal{L}}_n(\mathbf{X})\|_F^2 + \frac{1}{16}\|\mathbf{U}^\top\mathbf{U} - \mathbf{V}^\top\mathbf{V}\|_F^2$$
$$- \frac{4\bar{L}+1}{8}\|\mathbf{H}\|_F^4 - \left(\frac{r}{2\bar{L}} + \frac{2r}{\bar{\mu}}\right)\|\nabla\mathcal{L}_n(\mathbf{X}) - \nabla\bar{\mathcal{L}}_n(\mathbf{X})\|_2^2,$$

where $\mathbf{X} = \mathbf{U}\mathbf{V}^\top$, and $\bar{\mu}' = \min\{\bar{\mu}, 1\}$.

**Lemma A.2** (Local Smoothness Condition). For any $\mathbf{Z} = [\mathbf{U}; \mathbf{V}] \in \mathbb{R}^{(d_1+d_2)\times r}$, if we denote $\mathbf{X} = \mathbf{U}\mathbf{V}^\top$, then we have

$$\|\nabla\widetilde{F}_n(\mathbf{Z})\|_F^2 \leq (8\|\nabla\bar{\mathcal{L}}_n(\mathbf{X})\|_F^2 + \|\mathbf{U}^\top\mathbf{U} - \mathbf{V}^\top\mathbf{V}\|_F^2 + 8r\|\nabla\mathcal{L}_n(\mathbf{X}) - \nabla\bar{\mathcal{L}}_n(\mathbf{X})\|_2^2)\cdot\|\mathbf{Z}\|_2^2.$$

Now we are ready to prove Theorem 5.8.

*Proof of Theorem 5.8.* Let $\mathbf{Z}^t = [\mathbf{U}^t; \mathbf{V}^t]$, where $\mathbf{U}^t \in \mathbb{R}^{d_1 \times r}$ and $\mathbf{V}^t \in \mathbb{R}^{d_2 \times r}$. According to (A.1), the gradient descent based update in Algorithm 1 can be written as

$$\mathbf{Z}^{t+1} = \mathcal{P}_{\mathcal{C}}(\mathbf{Z}^t - \eta\nabla\widetilde{F}_n(\mathbf{Z}^t)), \tag{A.2}$$

where $\mathcal{C} \subseteq \mathbb{R}^{(d_1+d_2)\times r}$ is the corresponding rotation-invariant set based on $\mathcal{C}_1$ and $\mathcal{C}_2$ in Algorithm 1. To simplify notations, we denote the optimal rotation with respect to $\mathbf{Z}^t$ as $\mathbf{R}^t = \operatorname{argmin}_{\mathbf{R}\in\mathbb{Q}_r}\|\mathbf{Z}^t -$ $\mathbf{Z}^* \mathbf{R}\|_F$, and $\mathbf{H}^t = \mathbf{Z}^t - \mathbf{Z}^* \mathbf{R}^t$. According to assumption, by induction, we have $\mathbf{Z}^t \in \mathbb{B}(c_2 \sqrt{\sigma_r})$, for any $t \geq 0$. According to (A.2), we have

$$\begin{aligned} d^2(\mathbf{Z}^{t+1}, \mathbf{Z}^*) &\leq \|\mathbf{Z}^{t+1} - \mathbf{Z}^* \mathbf{R}^t\|_F^2 \\ &\leq \|\mathbf{Z}^t - \eta \nabla \widetilde{F}_n(\mathbf{Z}^t) - \mathbf{Z}^* \mathbf{R}^t\|_F^2 \\ &= d^2(\mathbf{Z}^t, \mathbf{Z}^*) - 2\eta \langle \nabla \widetilde{F}_n(\mathbf{Z}^t), \mathbf{H}^t \rangle + \eta^2 \|\nabla \widetilde{F}_n(\mathbf{Z}^t)\|_F^2, \end{aligned} \tag{A.3}$$

where the first inequality follows from Definition 5.1, and the second inequality follows from the non-expansion property of $\mathcal{P}_{\mathcal{C}}$, and the fact that $\mathbf{Z}^* \in \mathcal{C}$ and $\mathcal{C}$ is rotation-invariant. According to Lemma A.1, we have

$$\begin{aligned} \langle \nabla \widetilde{F}_n(\mathbf{Z}^t), \mathbf{H}^t \rangle &\geq \frac{\bar{\mu}' \sigma_r}{10} \|\mathbf{H}^t\|_F^2 + \frac{1}{4\bar{L}} \|\nabla \bar{\mathcal{L}}_n(\mathbf{X}^t)\|_F^2 + \frac{1}{16} \|\mathbf{U}^{t\top} \mathbf{U}^t - \mathbf{V}^{t\top} \mathbf{V}^t\|_F^2 \\ &\quad - \frac{4\bar{L}+1}{8} \|\mathbf{H}^t\|_F^4 - \left( \frac{r}{2\bar{L}} + \frac{2r}{\bar{\mu}} \right) \|\nabla \mathcal{L}_n(\mathbf{X}^t) - \nabla \bar{\mathcal{L}}_n(\mathbf{X}^t)\|_2^2. \end{aligned}$$

According to Lemma A.2, we have

$$\|\nabla \widetilde{F}_n(\mathbf{Z}^t)\|_F^2 \leq (8\|\nabla \bar{\mathcal{L}}_n(\mathbf{X}^t)\|_F^2 + \|\mathbf{U}^{t\top} \mathbf{U}^t - \mathbf{V}^{t\top} \mathbf{V}^t\|_F^2 + 8r\|\nabla \mathcal{L}_n(\mathbf{X}^t) - \nabla \bar{\mathcal{L}}_n(\mathbf{X}^t)\|_2^2) \cdot \|\mathbf{Z}^t\|_2^2.$$

Since $c_2 \leq 1/4$, we have $\|\mathbf{H}^t\|_F \leq \sqrt{\sigma_r}/4$, which implies $\|\mathbf{Z}^t\|_2 \leq \|\mathbf{Z}^* \mathbf{R}^t\|_2 + \|\mathbf{H}^t\|_2 \leq 2\sqrt{\sigma_1}$, due to the fact that $\|\mathbf{Z}^*\|_2^2 = 2\sigma_1$. Therefore, with $\eta = c_1/\sigma_1$, where $c_1 \leq \min\{1/(64\bar{L}), 1/32\}$, we have

$$-2\eta \langle \widetilde{F}_n(\mathbf{Z}^t), \mathbf{H}^t \rangle + \eta^2 \|\widetilde{F}_n(\mathbf{Z}^t)\|_F^2 \leq -\frac{\eta \bar{\mu}' \sigma_r}{5} \|\mathbf{H}^t\|_F^2 + \frac{\eta(4\bar{L}+1)}{4} \|\mathbf{H}^t\|_F^4 + c_3 \eta r \|\nabla \mathcal{L}_n(\mathbf{X}^t) - \nabla \bar{\mathcal{L}}_n(\mathbf{X}^t)\|_2^2,$$

where $c_3 = 2/\bar{L} + 4/\bar{\mu}$. Therefore, by inductive assumption $\|\mathbf{H}^t\|_F^2 \leq c_2^2 \sigma_r$, where $c_2^2 \leq 2\bar{\mu}'/\{5(4\bar{L}+1)\}$, we further obtain

$$-2\eta \langle \nabla \widetilde{F}_n(\mathbf{Z}^t), \mathbf{H}^t \rangle + \eta^2 \|\widetilde{F}_n(\mathbf{Z}^t)\|_F^2 \leq -\frac{\eta \bar{\mu}' \sigma_r}{10} \|\mathbf{H}^t\|_F^2 + c_3 \eta r \|\nabla \mathcal{L}_n(\mathbf{X}^t) - \nabla \bar{\mathcal{L}}_n(\mathbf{X}^t)\|_2^2. \tag{A.4}$$

Thus, according to Condition 5.7, by plugging (A.4) into (A.3), we have

$$d^2(\mathbf{Z}^{t+1}, \mathbf{Z}^*) \leq \left( 1 - \frac{\eta \bar{\mu}' \sigma_r}{10} \right) \|\mathbf{H}^t\|_F^2 + c_3 \eta r \epsilon^2(n, \delta),$$

holds with probability at least $1 - \delta$, which completes the proof. $\qquad \square$

## A.2 Proof of Theorem 5.9

While our proposed initialization algorithm is inspired by Jain et al. (2010), the proof of its theoretical guarantee is very different from that in Jain et al. (2010). Below, we show its detailed proof.

*Proof.* Consider the gradient descent based update in Algorithm 2, we have

$$\mathbf{X}_t = \mathcal{P}_r[\mathbf{X}_{t-1} - \tau \nabla \mathcal{L}_n(\mathbf{X}_{t-1})].$$

Recall $\mathbf{X}^* = \overline{\mathbf{U}}^* \mathbf{\Sigma}^* \overline{\mathbf{V}}^{*\top}$. For each iteration $t$, let the singular value decomposition of $\mathbf{X}_t$ be $\mathbf{X}_t = \mathbf{U}_t \mathbf{\Sigma}_t \mathbf{V}_t$. For any matrix $\mathbf{X}$, denote its row space and column space by $\text{row}(\mathbf{X})$ and $\text{col}(\mathbf{X})$, respectively. We define the following subspace spanned by the column vectors of $\overline{\mathbf{U}}^*$, $\mathbf{U}_{t-1}$ and $\mathbf{U}_t$ as

$$\text{span}(\mathbf{U}_{3r}) = \text{span}\{\overline{\mathbf{U}}^*, \mathbf{U}_{t-1}, \mathbf{U}_t\} = \text{col}(\overline{\mathbf{U}}^*) + \text{col}(\mathbf{U}_{t-1}) + \text{col}(\mathbf{U}_t),$$

where each column vector of $\mathbf{U}_{3r}$ is a basis vector of the above subspace, and the sum of two subspaces $\mathbf{U}_1, \mathbf{U}_2$ is defined as $\mathbf{U}_1 + \mathbf{U}_2 = \{\mathbf{u}_1 + \mathbf{u}_2 \mid \mathbf{u}_1 \in \mathbf{U}_1, \mathbf{u}_2 \in \mathbf{U}_2\}$. Similarly, we define the subspace spanned by the column vectors of $\overline{\mathbf{V}}^*$, $\mathbf{V}_{t-1}$ and $\mathbf{V}_t$ as

$$\text{span}(\mathbf{V}_{3r}) = \text{span}\{\overline{\mathbf{V}}^*, \mathbf{V}_{t-1}, \mathbf{V}_t\} = \text{col}(\overline{\mathbf{V}}^*) + \text{col}(\mathbf{V}_{t-1}) + \text{col}(\mathbf{V}_t),$$

Note that $\mathbf{X}^*, \mathbf{X}_{t-1}$ and $\mathbf{X}_t$ are all rank-$r$ matrices, thus both $\mathbf{U}_{3r}$ and $\mathbf{V}_{3r}$ have at most $3r$ columns. Moreover, we further define the following subspace

$$\mathcal{A}_{3r} = \{\boldsymbol{\Delta} \in \mathbb{R}^{d_1 \times d_2} \mid \text{row}(\boldsymbol{\Delta}) \subseteq \text{span}(\mathbf{V}_{3r}) \text{ and } \text{col}(\boldsymbol{\Delta}) \subseteq \text{span}(\mathbf{U}_{3r})\}.$$

Let $\Pi_{\mathcal{A}_{3r}}$ be the projection operator onto $\mathcal{A}_{3r}$. Specifically, for any $\mathbf{X} \in \mathbb{R}^{d_1 \times d_2}$, we have

$$\Pi_{\mathcal{A}_{3r}}(\mathbf{X}) = \mathbf{U}_{3r}\mathbf{U}_{3r}^\top \mathbf{X} \mathbf{V}_{3r}\mathbf{V}_{3r}^\top,$$

Note that for any $\mathbf{X} \in \mathbb{R}^{d_1 \times d_2}$, we have $\text{rank}\big(\Pi_{\mathcal{A}_{3r}}(\mathbf{X})\big) \leq 3r$, since $\text{rank}(\mathbf{A}\mathbf{B}) \leq \min\{\text{rank}(\mathbf{A}), \text{rank}(\mathbf{B})\}$. Besides, we denote

$$\widetilde{\mathbf{X}}_t = \mathbf{X}_{t-1} - \tau \Pi_{\mathcal{A}_{3r}}\big(\nabla \mathcal{L}_n(\mathbf{X}_{t-1})\big).$$

To begin with, we are going to upper bound $\|\widetilde{\mathbf{X}}_t - \mathbf{X}^*\|_F$. According to the triangle inequality, we have

$$
\begin{aligned}
\|\widetilde{\mathbf{X}}_t - \mathbf{X}^*\|_F &= \big\|\mathbf{X}_{t-1} - \mathbf{X}^* - \tau \Pi_{\mathcal{A}_{3r}}\big(\nabla \mathcal{L}_n(\mathbf{X}_{t-1})\big)\big\|_F \\
&\leq \big\|\mathbf{X}_{t-1} - \mathbf{X}^* - \tau \Pi_{\mathcal{A}_{3r}}\big(\nabla \mathcal{L}_n(\mathbf{X}_{t-1})\big) + \tau \Pi_{\mathcal{A}_{3r}}\big(\nabla \mathcal{L}_n(\mathbf{X}^*)\big)\big\|_F + \big\|\tau \Pi_{\mathcal{A}_{3r}}\big(\nabla \mathcal{L}_n(\mathbf{X}^*)\big)\big\|_F \\
&\leq \underbrace{\big\|\mathbf{X}_{t-1} - \mathbf{X}^* - \tau \big(\nabla \mathcal{L}_n(\mathbf{X}_{t-1}) - \nabla \mathcal{L}_n(\mathbf{X}^*)\big)\big\|_F}_{I_1} + \tau \underbrace{\big\|\Pi_{\mathcal{A}_{3r}}\big(\nabla \mathcal{L}_n(\mathbf{X}^*)\big)\big\|_F}_{I_2}, \quad\quad\text{(A.5)}
\end{aligned}
$$

where the first inequality follows from the triangle inequality, and the second inequality follows from the fact that $\mathbf{X}_{t-1} \in \mathcal{A}_{3r}$ and $\mathbf{X}^* \in \mathcal{A}_{3r}$, together with the non-expansion property of projection onto $\mathcal{A}_{3r}$. In the following discussions, we are going to bound $I_1$ and $I_2$ in (A.5), respectively. Consider $I_1$ first, we have

$$I_1^2 = \|\mathbf{X}_{t-1} - \mathbf{X}^*\|_F^2 - 2\tau \underbrace{\langle \mathbf{X}_{t-1} - \mathbf{X}^*, \nabla \mathcal{L}_n(\mathbf{X}_{t-1}) - \nabla \mathcal{L}_n(\mathbf{X}^*)\rangle}_{I_{11}} + \tau^2 \|\nabla \mathcal{L}_n(\mathbf{X}_{t-1}) - \nabla \mathcal{L}_n(\mathbf{X}^*)\|_F^2,$$

Note that $\mathcal{L}_n$ satisfies Conditions 5.3 and 5.4, which are parallel to Conditions 5.5 and 5.6. Thus according to Lemma C.2, we can use the same techniques to lower bound $I_{11}$

$$I_{11} \geq \frac{1}{2L}\|\nabla \mathcal{L}_n(\mathbf{X}_{t-1}) - \nabla \mathcal{L}_n(\mathbf{X}^*)\|_F^2 + \frac{\mu}{2}\|\mathbf{X}_{t-1} - \mathbf{X}^*\|_F^2,$$

If we set $\tau \leq 1/L$, then we obtain the upper bound of $I_1^2$

$$
\begin{aligned}
I_1^2 &\leq (1-\mu\tau)\|\mathbf{X}_{t-1} - \mathbf{X}^*\|_F^2 - \left(\frac{\tau}{L} - \tau^2\right)\|\nabla\mathcal{L}_n(\mathbf{X}_{t-1}) - \nabla\mathcal{L}_n(\mathbf{X}^*)\|_F^2 \\
&\leq (1-\mu\tau)\|\mathbf{X}_{t-1} - \mathbf{X}^*\|_F^2.
\end{aligned}
\tag{A.6}
$$

Next, we are going to upper bound $I_2$. We claim that for any matrix $\mathbf{X} \in \mathbb{R}^{d_1 \times d_2}$, we have $\|\Pi_{\mathcal{A}_{3r}}(\mathbf{X})\|_2 \leq \|\mathbf{X}\|_2$. In order to prove the claim, we are going to make use of the definition of operator norm. Accordingly, we have

$$
\begin{aligned}
\|\Pi_{\mathcal{A}_{3r}}(\mathbf{X})\|_2 &= \max_{\|\mathbf{u}\|_2 \leq 1, \|\mathbf{v}\|_2 \leq 1} \mathbf{u}^\top \Pi_{\mathcal{A}_{3r}}(\mathbf{X})\mathbf{v} \\
&= \max_{\|\mathbf{u}\|_2 \leq 1, \|\mathbf{v}\|_2 \leq 1} \mathbf{u}^\top \mathbf{U}_{3r}\mathbf{U}_{3r}^\top \mathbf{X} \mathbf{V}_{3r}\mathbf{V}_{3r}^\top \mathbf{v} \\
&\leq \max_{\|\mathbf{u}\|_2 \leq 1, \|\mathbf{v}\|_2 \leq 1} \mathbf{u}^\top \mathbf{X}\mathbf{v} = \|\mathbf{X}\|_2,
\end{aligned}
\tag{A.7}
$$

where the last inequality follows from the fact that $\|\mathbf{u}^\top \mathbf{U}_{3r}\mathbf{U}_{3r}^\top\|_2 \leq \|\mathbf{u}\|_2 \leq 1$ and $\|\mathbf{V}_{3r}\mathbf{V}_{3r}^\top \mathbf{v}\|_2 \leq \|\mathbf{v}\|_2 \leq 1$, because of the non-expansion property of projection. Note that $\text{rank}\big[\Pi_{\mathcal{A}_{3r}}(\nabla\mathcal{L}_n(\mathbf{X}^*))\big] \leq 3r$, then with probability at least $1-\delta$, we have

$$
I_2 \leq \sqrt{3r}\big\|\Pi_{\mathcal{A}_{3r}}(\nabla\mathcal{L}_n(\mathbf{X}^*))\big\|_2 \leq \sqrt{3r}\|\nabla\mathcal{L}_n(\mathbf{X}^*) - \nabla\bar{\mathcal{L}}_n(\mathbf{X}^*)\|_2 \leq \sqrt{3r}\epsilon(n,\delta),
\tag{A.8}
$$

where the second inequality follows from (A.7) and the fact that $\mathbf{X}^*$ minimize $\bar{\mathcal{L}}_n(\mathbf{X})$, and the last inequality follows from Condition 5.7. Finally, plugging (A.6) and (A.8) into (A.5), with probability at least $1-\delta$, we have

$$
\|\widetilde{\mathbf{X}}_t - \mathbf{X}^*\|_F \leq \sqrt{1-\mu\tau} \cdot \|\mathbf{X}_{t-1} - \mathbf{X}^*\|_F + \tau\sqrt{3r}\epsilon(n,\delta).
\tag{A.9}
$$

In the following discussions, we are going to show that $\mathbf{X}_t$ is the best rank-$r$ approximation of $\widetilde{\mathbf{X}}_t$. To simplify notations, denote $\mathbf{Y}_t = \mathbf{X}_{t-1} - \tau\nabla\mathcal{L}_n(\mathbf{X}_{t-1})$. Note that $\mathbf{X}_{t-1} \in \mathcal{A}_{3r}$ and the projection operator $\Pi_{\mathcal{A}_{3r}}$ is linear, then $\mathbf{X}_t = \mathcal{P}_r(\mathbf{Y}_t)$ and $\widetilde{\mathbf{X}}_t = \Pi_{\mathcal{A}_{3r}}(\mathbf{Y}_t)$. Let the singular value decomposition of $\mathbf{X}_t$ and $\mathbf{Y}_t$ be $\mathbf{X}_t = \mathbf{U}_1^t\mathbf{\Sigma}_1^t\mathbf{V}_1^{t\top}$, and $\mathbf{Y}_t = \mathbf{U}^t\mathbf{\Sigma}^t\mathbf{V}^{t\top} = \mathbf{U}_1^t\mathbf{\Sigma}_1^t\mathbf{V}_1^{t\top} + \mathbf{U}_2^t\mathbf{\Sigma}_2^t\mathbf{V}_2^{t\top}$, respectively. Then we have

$$
\widetilde{\mathbf{X}}_t = \Pi_{\mathcal{A}_{3r}}(\mathbf{U}_1^t\mathbf{\Sigma}_1^t\mathbf{V}_1^{t\top} + \mathbf{U}_2^t\mathbf{\Sigma}_2^t\mathbf{V}_2^{t\top}) = \mathbf{U}_1^t\mathbf{\Sigma}_1^t\mathbf{V}_1^{t\top} + \Pi_{\mathcal{A}_{3r}}(\mathbf{U}_2^t\mathbf{\Sigma}_2^t\mathbf{V}_2^{t\top}).
\tag{A.10}
$$

Let $d$ be the rank of $\mathbf{Y}_t$, and $\sigma_1^t \geq \sigma_2^t \geq \cdots \geq \sigma_d^t > 0$ be the sorted nonzero singular values of $\mathbf{Y}_t$. Since $\mathbf{X}_t = \mathcal{P}_r(\mathbf{Y}_t) = \mathbf{U}_1^t\mathbf{\Sigma}_1^t\mathbf{V}_1^{t\top}$, we have $\mathbf{\Sigma}_1^t = \text{diag}\{\sigma_1^t, \sigma_2^t, \cdots, \sigma_r^t\}$ and $\mathbf{\Sigma}_2^t = \text{diag}\{\sigma_{r+1}^t, \sigma_{r+2}^t, \cdots, \sigma_d^t\}$. Denote $\mathbf{W}_t = \Pi_{\mathcal{A}_{3r}}(\mathbf{U}_2^t\mathbf{\Sigma}_2^t\mathbf{V}_2^{t\top})$. Let the singular value decomposition of $\mathbf{W}_t$ be $\mathbf{W}_t = \widetilde{\mathbf{U}}_t\widetilde{\mathbf{\Sigma}}_t\widetilde{\mathbf{V}}_t^\top$, $\widetilde{r}$ be the rank of $\mathbf{W}_t$, and $\widetilde{\sigma}_1^t \geq \widetilde{\sigma}_2^t \geq \cdots \geq \widetilde{\sigma}_{\widetilde{r}}^t > 0$ be the sorted nonzero singular values of $\mathbf{W}_t$. Since $\text{col}(\mathbf{U}_1^t) \subseteq \text{span}(\mathbf{U}_{3r})$ and $\text{col}(\mathbf{U}_1^t) \perp \text{col}(\mathbf{U}_2^t)$, we have

$$
\mathbf{U}_1^{t\top}\mathbf{W}_t = \mathbf{U}_1^{t\top}\mathbf{U}_{3r}\mathbf{U}_{3r}^\top\mathbf{U}_2^t\mathbf{\Sigma}_2^t\mathbf{V}_2^{t\top}\mathbf{V}_{3r}\mathbf{V}_{3r}^\top = \mathbf{U}_1^{t\top}\mathbf{U}_2^t\mathbf{\Sigma}_2^t\mathbf{V}_2^{t\top}\mathbf{V}_{3r}\mathbf{V}_{3r}^\top = \mathbf{0},
$$

where the second equality holds because $\mathbf{U}_{3r}\mathbf{U}_{3r}^\top : \mathbb{R}^{d_1} \to \mathbb{R}^{d_1}$ can be regarded as a projection onto $\text{span}(\mathbf{U}_{3r})$, and every column vector of $\mathbf{U}_1^t$ belongs to $\text{span}(\mathbf{U}_{3r})$. Therefore, we conclude that $\mathbf{U}_1^{t\top}\widetilde{\mathbf{U}}_t = \mathbf{U}_1^{t\top}\mathbf{W}_t\widetilde{\mathbf{V}}_2^t(\widetilde{\mathbf{\Sigma}}_2^t)^{-1} = \mathbf{0}$, which implies that $\text{col}(\mathbf{U}_1^t) \perp \text{col}(\widetilde{\mathbf{U}}_t)$. Similarly, we have $\mathrm{col}(\mathbf{V}_1^t) \perp \mathrm{col}(\widetilde{\mathbf{V}}_t)$. Thus, according to (A.10), we actually obtain the singular value decomposition of $\widetilde{\mathbf{X}}_t$

$$\widetilde{\mathbf{X}}_t = \mathbf{U}_1^t \mathbf{\Sigma}_1^t \mathbf{V}_1^{t\top} + \widetilde{\mathbf{U}}_t \widetilde{\mathbf{\Sigma}}_t \widetilde{\mathbf{V}}_t^\top = \begin{bmatrix} \mathbf{U}_1^t & \widetilde{\mathbf{U}}_t \end{bmatrix} \begin{bmatrix} \mathbf{\Sigma}_1^t & \mathbf{0} \\ \mathbf{0} & \widetilde{\mathbf{\Sigma}}_t \end{bmatrix} \begin{bmatrix} \mathbf{V}_1^{t\top} \\ \widetilde{\mathbf{V}}_t^\top \end{bmatrix}.$$

According to (A.7), we have $\|\Pi_{\mathcal{A}_{3r}}(\mathbf{U}_2^t \mathbf{\Sigma}_2^t \mathbf{V}_2^{t\top})\|_2 \le \|\mathbf{U}_2^t \mathbf{\Sigma}_2^t \mathbf{V}_2^{t\top}\|_2$, which implies $\widetilde{\sigma}_1^t \le \sigma_{r+1}^t$. Therefore, we have $\mathcal{P}_r(\widetilde{\mathbf{X}}_t) = \mathbf{U}_1^t \mathbf{\Sigma}_1^t \mathbf{V}_1^t = \mathbf{X}_t$, which implies

$$\|\mathbf{X}_t - \mathbf{X}^*\|_F \le \|\mathbf{X}_t - \widetilde{\mathbf{X}}_t\|_F + \|\widetilde{\mathbf{X}}_t - \mathbf{X}^*\|_F \le 2\|\widetilde{\mathbf{X}}_t - \mathbf{X}^*\|_F, \tag{A.11}$$

where the first inequality follows from the triangle inequality, and the second inequality holds because $\mathbf{X}_t$ is the best rank-$r$ approximation of $\widetilde{\mathbf{X}}_t$, such that $\mathbf{X}_t = \mathrm{argmin}_{\mathrm{rank}(\mathbf{X}) \le r} \|\widetilde{\mathbf{X}}_t - \mathbf{X}\|_F$. Finally, by plugging (A.9) into (A.11), with probability at least $1 - \delta$, we have

$$\|\mathbf{X}_t - \mathbf{X}^*\|_F \le 2\sqrt{1 - \mu\tau} \cdot \|\mathbf{X}_{t-1} - \mathbf{X}^*\|_F + 2\tau\sqrt{3r}\epsilon(n, \delta).$$

Let the contraction parameter $\rho = 2\sqrt{1 - \mu\tau} < 1$, then it is sufficient to set $\tau > 3/(4\mu)$. Hence, with probability at least $1 - \delta$, we have

$$\|\mathbf{X}_t - \mathbf{X}^*\|_F \le \rho^t \|\mathbf{X}^*\|_F + \frac{2\tau\sqrt{3r}\epsilon(n, \delta)}{1 - \rho},$$

which completes the proof. □

# B Proofs of Specific Examples

## B.1 Proof of Corollary 5.11

We consider a more general model for observation matrix $\mathbf{A}_i$ with dependent entries. Denote $\mathrm{vec}(\mathbf{A}_i) \in \mathbb{R}^{d_1 d_2}$ as the vectorization of $\mathbf{A}_i$. Given a symmetric positive definite matrix $\mathbf{\Sigma} \in \mathbb{R}^{d_1 d_2 \times d_1 d_2}$, we say the observation matrix $\mathbf{A}_i$ is sampled from $\mathbf{\Sigma}$-ensemble, if $\mathrm{vec}(\mathbf{A}_i) \sim N(0, \mathbf{\Sigma})$ (Negahban and Wainwright, 2011). Define $\pi^2(\mathbf{\Sigma}) = \sup_{\|\mathbf{u}\|_2=1, \|\mathbf{v}\|_2=1} \mathrm{Var}(\mathbf{u}^\top \mathbf{A} \mathbf{v})$, where $\mathbf{A} \in \mathbb{R}^{d_1 \times d_2}$ is a random matrix sampled from $\mathbf{\Sigma}$-ensemble. Specifically, when $\mathbf{\Sigma} = \mathbf{I}$, it corresponds to the classical matrix regression model where the entries of $\mathbf{A}_i$ are independent from each other. Besides, in this case, we have $\pi(\mathbf{I}) = 1$.

For any matrix $\mathbf{X} \in \mathbb{R}^{d_1 \times d_2}$, recall the linear operator $\mathcal{A}(\mathbf{X}) = (\langle \mathbf{A}_1, \mathbf{X} \rangle, \langle \mathbf{A}_2, \mathbf{X} \rangle, \ldots, \langle \mathbf{A}_n, \mathbf{X} \rangle)^\top$. In order to establish the restricted strongly convexity and smoothness conditions, we need to make use of the following lemma, which has been used in Agarwal et al. (2010); Negahban and Wainwright (2011).

**Lemma B.1.** Consider the linear operator $\mathcal{A}$ with each element $\mathbf{A}_i$ sampled from $\mathbf{\Sigma}$-ensemble, then for all $\mathbf{\Delta} \in \mathbb{R}^{d_1 \times d_2}$, it holds with probability at least $1 - \exp(-C_0 n)$ that

$$\frac{\|\mathcal{A}(\mathbf{\Delta})\|_2^2}{n} \ge \frac{1}{2} \|\sqrt{\mathbf{\Sigma}} \mathrm{vec}(\mathbf{\Delta})\|_2^2 - C_1 \pi^2(\mathbf{\Sigma}) \frac{d'}{n} \|\mathbf{\Delta}\|_*^2, \tag{B.1}$$

$$\frac{\|\mathcal{A}(\mathbf{\Delta})\|_2^2}{n} \le \frac{1}{2} \|\sqrt{\mathbf{\Sigma}} \mathrm{vec}(\mathbf{\Delta})\|_2^2 + C_1 \pi^2(\mathbf{\Sigma}) \frac{d'}{n} \|\mathbf{\Delta}\|_*^2. \tag{B.2}$$

The following lemma, used in Negahban and Wainwright (2011), upper bounds the difference between the gradient of the sample loss function $\nabla \mathcal{L}_n$ and the gradient of the expected loss function $\nabla \bar{\mathcal{L}}_n$, as long as the noise satisfies $\|\epsilon\|_2 \leq 2\nu\sqrt{n}$ for some constant $\nu$. Obviously, this condition holds for any bounded noise, and it is proved in Vershynin (2010) that for any sub-Gaussian random noise with parameter $\nu$, this condition holds with high probability.

**Lemma B.2.** Consider the linear operator $\mathcal{A}$ with each element $\mathbf{A}_i$ sampled from $\mathbf{\Sigma}$-ensemble, and the noise vector $\epsilon$ satisfies that $\|\epsilon\|_2 \leq 2\nu\sqrt{n}$. Then there exist constants $C, C_1$ and $C_2$, such that with probability at least $1 - C_1\exp(-C_2 d')$ we have

$$\left\| \frac{1}{n}\sum_{i=1}^n \epsilon_i \mathbf{A}_i \right\|_2 \leq C\nu\sqrt{\frac{d'}{n}}.$$

*Proof of Corollary 5.11.* First, we are going to prove the restricted strongly convexity condition for $\mathcal{L}_n$ and $\bar{\mathcal{L}}_n$. Since we have

$$\mathcal{L}_n(\mathbf{X}) = \frac{1}{2n}\sum_{i=1}^n \left(\langle \mathbf{A}_i, \mathbf{X}\rangle - y_i\right)^2 = \frac{1}{2n}\sum_{i=1}^n \left(\langle \mathbf{A}_i, \mathbf{X} - \mathbf{X}^*\rangle - \epsilon_i\right)^2,$$

thus by taking expectation with respect to noise, we obtain the expression of $\bar{\mathcal{L}}_n$

$$\bar{\mathcal{L}}_n(\mathbf{X}) = \mathbb{E}[\mathcal{L}_n(\mathbf{X})] = \frac{1}{2n}\sum_{i=1}^n \left(\langle \mathbf{A}_i, \mathbf{X} - \mathbf{X}^*\rangle\right)^2 + \frac{1}{2}\mathbb{E}(\epsilon_i^2).$$

Let $\mathbf{X}, \mathbf{Y} \in \mathbb{R}^{d_1 \times d_2}$ be any two rank-$r$ matrices, then for the sample loss function $\mathcal{L}_n$, we have

$$\begin{aligned}
\mathcal{L}_n(\mathbf{Y}) - \mathcal{L}_n(\mathbf{X}) - \langle \nabla\mathcal{L}_n(\mathbf{X}), \mathbf{\Delta}\rangle &= \frac{1}{2n}\sum_{i=1}^n \left(\langle \mathbf{A}_i, \mathbf{Y} - \mathbf{X}^*\rangle^2 - \langle \mathbf{A}_i, \mathbf{X} - \mathbf{X}^*\rangle^2 - 2\langle \mathbf{A}_i, \mathbf{X} - \mathbf{X}^*\rangle\langle \mathbf{A}_i, \mathbf{\Delta}\rangle\right) \\
&= \frac{\|\mathcal{A}(\mathbf{\Delta})\|_2^2}{2n},
\end{aligned} \tag{B.3}$$

where $\mathbf{\Delta} = \mathbf{Y} - \mathbf{X}$. Similarly, for the expected loss function $\bar{\mathcal{L}}_n$, we have

$$\begin{aligned}
\bar{\mathcal{L}}_n(\mathbf{Y}) - \bar{\mathcal{L}}_n(\mathbf{X}) - \langle \nabla\bar{\mathcal{L}}_n(\mathbf{X}), \mathbf{\Delta}\rangle &= \frac{1}{2n}\sum_{i=1}^n \left(\langle \mathbf{A}_i, \mathbf{Y} - \mathbf{X}^*\rangle^2 - \langle \mathbf{A}_i, \mathbf{X} - \mathbf{X}^*\rangle^2 - 2\langle \mathbf{A}_i, \mathbf{X} - \mathbf{X}^*\rangle\langle \mathbf{A}_i, \mathbf{\Delta}\rangle\right) \\
&= \frac{\|\mathcal{A}(\mathbf{\Delta})\|_2^2}{2n}.
\end{aligned} \tag{B.4}$$

Thus, according to (B.3) and (B.4), it is sufficient to bound the term $\|\mathcal{A}(\mathbf{\Delta})\|_2^2/n$ for both $\mathcal{L}_n$ and $\bar{\mathcal{L}}_n$. According to (B.1) in Lemma B.1, we have

$$\frac{\|\mathcal{A}(\mathbf{\Delta})\|_2^2}{n} \geq \frac{1}{2}\left\|\sqrt{\mathbf{\Sigma}}\text{vec}(\mathbf{\Delta})\right\|_2^2 - C_1\pi^2(\mathbf{\Sigma})\frac{d'}{n}\|\mathbf{\Delta}\|_*^2.$$

Since both $\mathbf{X}$ and $\mathbf{Y}$ are rank-$r$ matrices, we have rank$(\mathbf{\Delta}) \leq 2r$, which implies that $\|\mathbf{\Delta}\|_* \leq \sqrt{2r}\|\mathbf{\Delta}\|_F$. Therefore, we have

$$\frac{\|\mathcal{A}(\mathbf{\Delta})\|_2^2}{n} \geq \left\{\frac{\lambda_{\min}(\mathbf{\Sigma})}{2} - 2C_1 r\pi^2(\mathbf{\Sigma})\frac{d'}{n}\right\}\|\mathbf{\Delta}\|_F^2.$$

Thus, for $n \geq C_3 \pi^2(\mathbf{\Sigma}) r d' / \lambda_{\min}(\mathbf{\Sigma})$, where $C_3$ is sufficiently large such that

$$2C_1 r \pi^2(\mathbf{\Sigma}) \frac{d'}{n} \leq \frac{\lambda_{\min}(\mathbf{\Sigma})}{18},$$

we have

$$\frac{\|\mathcal{A}(\mathbf{\Delta})\|_2^2}{n} \geq \frac{4\lambda_{\min}(\mathbf{\Sigma})}{9} \|\mathbf{\Delta}\|_F^2.$$

Since $\mathbf{\Sigma} = \mathbf{I}$ under our matrix regression model, we obtain $\mu = 4/9$. Next, we prove the restricted strongly smoothness condition for $\mathcal{L}_n$ and $\bar{\mathcal{L}}_n$. According to (B.2) in Lemma B.1, we have

$$\frac{\|\mathcal{A}(\mathbf{\Delta})\|_2^2}{n} \leq \frac{1}{2} \|\sqrt{\mathbf{\Sigma}} \text{vec}(\mathbf{\Delta})\|_2^2 + C_1 \pi^2(\mathbf{\Sigma}) \frac{d'}{n} \|\mathbf{\Delta}\|_*^2.$$

Similarly, we have $\text{rank}(\mathbf{\Delta}) \leq 2r$, which implies that $\|\mathbf{\Delta}\|_* \leq \sqrt{2r} \|\mathbf{\Delta}\|_F$. Therefore, we can get

$$\frac{\|\mathcal{A}(\mathbf{\Delta})\|_2^2}{n} \leq \left\{ \frac{\lambda_{\max}(\mathbf{\Sigma})}{2} + 2C_1 r \pi^2(\mathbf{\Sigma}) \frac{d'}{n} \right\} \|\mathbf{\Delta}\|_F^2.$$

Thus, for $n \geq C_3 \pi^2(\mathbf{\Sigma}) r d' / \lambda_{\min}(\mathbf{\Sigma})$, where $C_3$ is sufficiently large such that

$$2C_1 r \pi^2(\mathbf{\Sigma}) \frac{d'}{n} \leq \frac{\lambda_{\min}(\mathbf{\Sigma})}{18},$$

we have

$$\frac{\|\mathcal{A}(\mathbf{\Delta})\|_2^2}{n} \leq \frac{5\lambda_{\max}(\mathbf{\Sigma})}{9} \|\mathbf{\Delta}\|_F^2.$$

Therefore, due to the fact that $\mathbf{\Sigma} = \mathbf{I}$, we have $L = 5/9$. Finally, we upper bound the term $\|\nabla \mathcal{L}_n(\mathbf{X}) - \nabla \bar{\mathcal{L}}_n(\mathbf{X})\|_2^2$. By the definition of $\mathcal{L}_n$ and $\bar{\mathcal{L}}_n$, we have

$$\nabla \bar{\mathcal{L}}_n(\mathbf{X}) - \nabla \mathcal{L}_n(\mathbf{X}) = \frac{1}{n} \sum_{i=1}^n \epsilon_i \mathbf{A}_i.$$

According to Lemma B.2, there exist constants $C, C_1'$ and $C_2'$ such that

$$\left\| \frac{1}{n} \sum_{i=1}^n \epsilon_i \mathbf{A}_i \right\|_2 \leq C \nu \sqrt{\frac{d'}{n}},$$

hold with probability at least $1 - C_1' \exp(-C_2' d')$. Thus, we obtain

$$\|\nabla \mathcal{L}_n(\mathbf{X}) - \nabla \bar{\mathcal{L}}_n(\mathbf{X})\|_2^2 \leq C^2 \nu^2 \frac{d'}{n},$$

hold with probability at least $1 - C_1' \exp(-C_2' d')$. Therefore, by applying Theorem 5.8, we obtain the linear convergence results for matrix regression. $\qquad \square$

## B.2 Proof of Corollary 5.13

As discussed in Gross (2011) that if $\mathbf{X}^*$ is equal to zero in nearly all of rows or columns, then it is impossible to recovery $\mathbf{X}^*$ unless all of its entries are sampled. In other words, there will always be some low-rank matrices which are too spiky to be recovered without sampling the whole matrix. In order to avoid the overly spiky matrices in matrix completion, existing work (Candès and Recht, 2009) imposes stringent matrix incoherence conditions to preclude such matrices. These constraints are relaxed in more recent work (Negahban and Wainwright, 2012) by restricting the spikiness ratio, which is defined as follows: $\alpha_{\mathrm{sp}}(\mathbf{X}) = (\sqrt{d_1 d_2} \|\mathbf{X}\|_\infty)/\|\mathbf{X}\|_F$. Therefore, we consider the following spikiness constraint $\|\mathbf{X}\|_\infty \le \alpha$, where $\alpha = \alpha_{\mathrm{sp}}(\mathbf{X}) \|\mathbf{X}\|_F / \sqrt{d_1 d_2}$.

We obtain the restricted strongly convexity and smoothness conditions for both $\mathcal{L}_n$ and $\bar{\mathcal{L}}_n$, with parameters $\mu = 8/9$ and $L = 10/9$ under the spikiness constraint. In order to establish the restricted strongly convexity and smoothness conditions, we need to make use of the following lemma, which comes from Negahban and Wainwright (2012).

**Lemma B.3.** There exist universal constants $c_1, c_2, c_3, c_4, c_5$ such that as long as $n > c_1 d' \log d'$, if the following condition is satisfied

$$\alpha_{\mathrm{sp}}(\boldsymbol{\Delta}) \frac{\|\boldsymbol{\Delta}\|_*}{\|\boldsymbol{\Delta}\|_F} \le \frac{1}{c_2} \sqrt{n/d' \log d'}, \tag{B.5}$$

we have, with probability at least $1 - c_3 \exp(-c_4 d' \log d')$, that the following holds

$$\left| \frac{\|\mathcal{A}(\boldsymbol{\Delta})\|_2}{\sqrt{n}} - \frac{\|\boldsymbol{\Delta}\|_F}{\sqrt{d_1 d_2}} \right| \le \frac{1}{10} \frac{\|\boldsymbol{\Delta}\|_F}{\sqrt{d_1 d_2}} \left( 1 + \frac{c_5 \alpha_{\mathrm{sp}}(\boldsymbol{\Delta})}{\sqrt{n}} \right).$$

For matrix completion, we have the following matrix concentration inequality, which establishes the upper bound of the difference between the gradient of the sample loss function $\nabla \mathcal{L}_n$ and the gradient of the expected loss function $\nabla \bar{\mathcal{L}}_n$.

**Lemma B.4.** (Negahban and Wainwright, 2012) Let $\mathbf{A}_i$ be uniformly distributed on $\mathcal{X}$ and each $\epsilon_i$ be i.i.d. zero mean Gaussian variable with variance $\nu^2$. Then, there exist constants $c_1$ and $c_2$ such that with probability at least $1 - c_1/d'$ we have

$$\left\| \frac{1}{n} \sum_{i=1}^n \epsilon_i \mathbf{A}_i \right\|_2 \le c_2 \nu \sqrt{\frac{d' \log d'}{d_1 d_2 n}}.$$

*Proof of Corollary 5.13.* Since matrix completion is a special case of matrix regression, the convexity and smoothness properties of $\mathcal{L}_n$ and $\bar{\mathcal{L}}_n$ are the same, thus we only need to verify the restricted strongly convexity and smoothness conditions of $\bar{\mathcal{L}}_n$. In order to prove our results, we are going to consider the following two cases. Recall $\mathcal{C}(\alpha) = \left\{ \mathbf{X} \in \mathbb{R}^{d_1 \times d_2} : \|\mathbf{X}\|_\infty \le \alpha \right\}$. Let $\mathbf{X}, \mathbf{Y} \in \mathcal{C}(\alpha)$ be any two rank-$r$ matrices, denote $\boldsymbol{\Delta} = \mathbf{Y} - \mathbf{X}$.

**Case 1:** Suppose condition (B.5) is not satisfied. Then we have

$$\|\boldsymbol{\Delta}\|_F^2 \le C_0 \left( \sqrt{d_1 d_2} \|\boldsymbol{\Delta}\|_\infty \right) \|\boldsymbol{\Delta}\|_* \sqrt{\frac{d' \log d'}{n}}$$

$$\le 2 C_0 \alpha \sqrt{d_1 d_2} \|\boldsymbol{\Delta}\|_* \sqrt{\frac{d' \log d'}{n}},$$

where the second inequality holds because $\|\mathbf{\Delta}\|_\infty \leq \|\mathbf{X}\|_\infty + \|\mathbf{Y}\|_\infty \leq 2\alpha$. Note that we have $\mathrm{rank}(\mathbf{\Delta}) \leq 2r$, we can obtain

$$\|\mathbf{\Delta}\|_F^2 \leq 2C_0\alpha\sqrt{2r}\sqrt{d_1 d_2}\|\mathbf{\Delta}\|_F\sqrt{\frac{d'\log d'}{n}},$$

which implies that

$$\frac{1}{d_1 d_2}\|\mathbf{\Delta}\|_F^2 \leq C\alpha^2 \frac{rd'\log d'}{n}. \tag{B.6}$$

**Case** 2: Suppose condition (B.5) is satisfied. First, we prove the restricted strongly convexity condition of the expected loss function $\bar{\mathcal{L}}_n$ with parameter $\mu = 8/9$. We have

$$\begin{aligned}
\bar{\mathcal{L}}_n(\mathbf{Y}) - \bar{\mathcal{L}}_n(\mathbf{X}) - \langle \nabla\bar{\mathcal{L}}_n(\mathbf{X}), \mathbf{\Delta}\rangle &= \frac{1}{2p}\sum_{i=1}^n \left(\langle\mathbf{A}_i, \mathbf{Y} - \mathbf{X}^*\rangle + \langle\mathbf{A}_i, \mathbf{X} - \mathbf{X}^*\rangle^2 - 2\langle\mathbf{A}_i, \mathbf{X} - \mathbf{X}^*\rangle\langle\mathbf{A}_i, \mathbf{\Delta}\rangle\right. \\
&= \frac{\|\mathcal{A}(\mathbf{\Delta})\|_2^2}{2p}, \tag{B.7}
\end{aligned}$$

Therefore, if $c_5\alpha_{\mathrm{sp}}(\mathbf{\Delta})/\sqrt{n} \geq 1/9$, according to the definition of $\alpha_{\mathrm{sp}}(\mathbf{\Delta})$, we can obtain

$$\|\mathbf{\Delta}\|_F \leq c'\alpha\sqrt{\frac{d_1 d_2}{n}},$$

where $c'$ is a constant, which implies that

$$\frac{1}{d_1 d_2}\|\mathbf{\Delta}\|_F^2 \leq c'\alpha^2\frac{1}{n}, \tag{B.8}$$

On the other hand, if $c_5\alpha_{\mathrm{sp}}(\mathbf{\Delta})/\sqrt{n} \leq 1/9$, by Lemma B.3, we have

$$\frac{\|\mathcal{A}(\mathbf{\Delta})\|_2^2}{p} \geq \frac{8}{9}\|\mathbf{\Delta}\|_F^2,$$

and thus we conclude $\mu = 8/9$. Next, we prove the restrict strongly smoothness condition for the expected loss function $\bar{\mathcal{L}}_n$ with parameter $L = 10/9$. Since we have

$$\bar{\mathcal{L}}_n(\mathbf{Y}) - \bar{\mathcal{L}}_n(\mathbf{X}) - \langle\nabla\bar{\mathcal{L}}_n(\mathbf{X}), \mathbf{\Delta}\rangle = \frac{\|\mathcal{A}(\mathbf{\Delta})\|_2^2}{2p},$$

according to Lemma B.3, as long as $c_5\alpha_{\mathrm{sp}}(\Delta)/\sqrt{n} \leq 1/9$, we have

$$\frac{\|\mathcal{A}(\mathbf{\Delta})\|_2^2}{p} \leq \frac{10}{9}\|\mathbf{\Delta}\|_F^2,$$

and thus we conclude $L = 10/9$. Finally, we bound the term $\|\nabla\mathcal{L}_n(\mathbf{X}) - \nabla\bar{\mathcal{L}}_n(\mathbf{X})\|_2^2$. By the definition of $\mathcal{L}_n$ and $\bar{\mathcal{L}}_n$, we have

$$\nabla\bar{\mathcal{L}}_n(\mathbf{X}) - \nabla\mathcal{L}_n(\mathbf{X}) = \frac{1}{p}\sum_{i=1}^n \epsilon_i\mathbf{A}_i.$$

Recall that the noise matrix has i.i.d. entries such that each entry follows sub-Gaussian distribution with parameter $\nu$. Thus, according to Lemma B.4, we have that

$$\Big\| \frac{1}{p} \sum_{i=1}^{n} \epsilon_i \mathbf{A}_i \Big\|_2 \leq C\nu \sqrt{d_1 d_2} \sqrt{\frac{d' \log d'}{n}},$$

holds with probability at least $1 - C'/d'$. Thus, we obtain

$$\|\nabla \mathcal{L}_n(\mathbf{X}) - \nabla \bar{\mathcal{L}}_n(\mathbf{X})\|_2^2 \leq C^2 \nu^2 d_1 d_2 \frac{d' \log d'}{n}, \tag{B.9}$$

holds with probability at least $1 - C'/d'$. Since the error bound (B.6), (B.8) and (B.9) are dominated by the following statistical error bound

$$C_1 \max\{\nu^2, \alpha^2\} \frac{d' \log d'}{p},$$

where $C_1$ is a universal constant, which completes our proof. $\qquad \square$

## B.3   Proof of Corollary 5.15

We are going to obtain the restricted strongly convexity and smoothness conditions for both $\mathcal{L}_n$ and $\bar{\mathcal{L}}_n$ under spikiness condition, with parameters $C_1 \mu_\alpha$ and $C_2 L_\alpha$, respectively, where $\mu_\alpha$ and $L_\alpha$ are defined according to (5.1) and (5.2). In order to bound the error between the gradient of $\mathcal{L}_n$ and $\bar{\mathcal{L}}_n$, we need to make use of the following lemma.

**Lemma B.5.** (Negahban and Wainwright, 2012) If $\Omega \subseteq [d_1] \times [d_2]$ is sampled under uniform model, then we have the following upper bound

$$\Big\| \frac{1}{n} \sum_{i=1}^{n} \sqrt{d_1 d_2} \mathbf{e}_{j(d_1)}^{i} \mathbf{e}_{k(d_2)}^{i\top} \Big\|_2 \leq C\sqrt{\frac{d' \log d'}{n}},$$

holds with probability at least $1 - C'/d$, where $C, C'$ are universal constants.

*Proof of Corollary 5.15.* Let $|\Omega| = n$, $\mathbf{A}_i = \mathbf{e}_{j(d1)}^{i} \mathbf{e}_{k(d2)}^{i\top}$, we can rewrite the sample loss function as follows

$$\mathcal{L}_n(\mathbf{X}) := -\frac{1}{p} \sum_{i=1}^{n} \big\{ \mathbb{1}_{(y_i=1)} \log \big( f(\langle \mathbf{A}_i, \mathbf{X} \rangle) \big) + \mathbb{1}_{(y_i=-1)} \log \big( 1 - f(\langle \mathbf{A}_i, \mathbf{X} \rangle) \big) \big\},$$

where $p = n/d_1 d_2$. Therefore, we have the expected loss function $\bar{\mathcal{L}}_n$ as

$$\bar{\mathcal{L}}_n(\mathbf{X}) := -\frac{1}{p} \sum_{i=1}^{n} \big\{ f(\langle \mathbf{A}_i, \mathbf{X}^* \rangle) \log \big( f(\langle \mathbf{A}_i, \mathbf{X} \rangle) \big) + \big( 1 - f(\langle \mathbf{A}_i, \mathbf{X}^* \rangle) \big) \log \big( 1 - f(\langle \mathbf{A}_i, \mathbf{X} \rangle) \big) \big\}, \tag{B.10}$$

which implies

$$\nabla \bar{\mathcal{L}}_n(\mathbf{X}) = \frac{1}{p} \sum_{i=1}^{n} \bigg( -\frac{f'(\langle \mathbf{A}_i, \mathbf{X} \rangle)}{f(\langle \mathbf{A}_i, \mathbf{X} \rangle)} f(\langle \mathbf{A}_i, \mathbf{X}^* \rangle) + \frac{f'(\langle \mathbf{A}_i, \mathbf{X} \rangle)}{1 - f(\langle \mathbf{A}_i, \mathbf{X} \rangle)} \big( 1 - f(\langle \mathbf{A}_i, \mathbf{X}^* \rangle) \big) \bigg) \mathbf{A}_i. \tag{B.11}$$

Furthermore, we obtain

$$\nabla^2 \bar{\mathcal{L}}_n(\mathbf{X}) = \frac{1}{p} \sum_{i=1}^n B_i(\mathbf{X}) \text{vec}(\mathbf{A}_i) \text{vec}(\mathbf{A}_i)^\top,$$

where we have

$$\begin{aligned}
B_i(\mathbf{X}) = \Bigg[ &\left( \frac{f'^2(\langle \mathbf{A}_i, \mathbf{X} \rangle)}{f^2(\langle \mathbf{A}_i, \mathbf{X} \rangle)} - \frac{f''(\langle \mathbf{A}_i, \mathbf{X} \rangle)}{f(\langle \mathbf{A}_i, \mathbf{X} \rangle)} \right) f(\langle \mathbf{A}_i, \mathbf{X}^* \rangle) \\
&+ \left( \frac{f''(\langle \mathbf{A}_i, \mathbf{X} \rangle)}{1 - f(\langle \mathbf{A}_i, \mathbf{X} \rangle)} - \frac{f'^2(\langle \mathbf{A}_i, \mathbf{X} \rangle)}{(1 - f(\langle \mathbf{A}_i, \mathbf{X} \rangle)^2)} \right) \big( 1 - f(\langle \mathbf{A}_i, \mathbf{X}^* \rangle) \big) \Bigg].
\end{aligned}$$

Next, we prove the strongly convexity and smoothness condition by the mean value theorem. Since for any $\mathbf{X}, \mathbf{M} \in \mathbb{R}^{d_1 \times d_2}$, we have

$$\bar{\mathcal{L}}_n(\mathbf{X}) = \bar{\mathcal{L}}_n(\mathbf{M}) + \langle \nabla \bar{\mathcal{L}}_n(\mathbf{M}), \mathbf{X} - \mathbf{M} \rangle + \frac{1}{2} (\mathbf{x} - \mathbf{m})^\top \nabla^2 \bar{\mathcal{L}}_n(\mathbf{W})(\mathbf{x} - \mathbf{m}),$$

where we have $\mathbf{W} = \mathbf{M} + a(\mathbf{X} - \mathbf{M})$ for $a \in [0, 1]$, and $\mathbf{x} = \text{vec}(\mathbf{X})$, $\mathbf{m} = \text{vec}(\mathbf{M})$. Furthermore, we have

$$\begin{aligned}
(\mathbf{x} - \mathbf{m})^\top \nabla^2 \bar{\mathcal{L}}_n(\mathbf{W})(\mathbf{x} - \mathbf{m}) &= \frac{1}{p} \sum_{i=1}^n B_i(\mathbf{W}) \langle \text{vec}(\mathbf{A}_i)^\top (\mathbf{x} - \mathbf{m}), \text{vec}(\mathbf{A}_i)^\top (\mathbf{x} - \mathbf{m}) \rangle \\
&= \frac{1}{p} \sum_{i=1}^n B_i(\mathbf{W}) \langle \mathbf{A}_i, \boldsymbol{\Delta} \rangle^2,
\end{aligned}$$

where $\boldsymbol{\Delta} = \mathbf{X} - \mathbf{M}$. Therefore, we can get

$$\frac{1}{p} \sum_{i=1}^n B_i(\mathbf{W}) \langle \mathbf{A}_i, \boldsymbol{\Delta} \rangle^2 \geq \mu_\alpha \frac{\|\mathcal{A}(\boldsymbol{\Delta})\|_2^2}{p},$$

where the inequality follows from the definition of $\mu_\alpha$ in (5.1). Thus, by the same proof as in the case of matrix completion, we can get

$$\mu_\alpha \frac{\|\mathcal{A}(\boldsymbol{\Delta})\|_2^2}{p} \geq C_1 \mu_\alpha \|\boldsymbol{\Delta}\|_F^2.$$

And thus we have $\mu = C_1 \mu_\alpha$. Therefore, we have

$$\bar{\mathcal{L}}_n(\mathbf{X}) \geq \bar{\mathcal{L}}_n(\mathbf{M}) + \langle \nabla \bar{\mathcal{L}}_n(\mathbf{M}), \mathbf{X} - \mathbf{M} \rangle + \frac{1}{2} C_1 \mu_\alpha \|\boldsymbol{\Delta}\|_F^2,$$

On the other hand, we have

$$\frac{1}{p} \sum_{i=1}^n B_i(\mathbf{W}) \langle \mathbf{A}_i, \boldsymbol{\Delta} \rangle^2 \leq L_\alpha \frac{\|\mathcal{A}(\boldsymbol{\Delta})\|_2^2}{p},$$

where the inequality follows from the definition of $L_\alpha$ in (5.2). Thus, by the same proof as matrix completion, we can get

$$L_\alpha \frac{\|\mathcal{A}(\boldsymbol{\Delta})\|_2^2}{p} \leq C_2 L_\alpha \|\boldsymbol{\Delta}\|_F^2.$$

Therefore, we have

$$\bar{\mathcal{L}}_n(\mathbf{X}) \leq \bar{\mathcal{L}}_n(\mathbf{M}) + \langle \nabla \bar{\mathcal{L}}_n(\mathbf{M}), \mathbf{X} - \mathbf{M} \rangle + \frac{1}{2} C_2 L_\alpha \|\boldsymbol{\Delta}\|_F^2,$$

where $L = C_2 L_\alpha$. Since the proof of the restricted strongly convexity and smoothness property of $\mathcal{L}_n$ is almost the same as $\bar{\mathcal{L}}_n$, we skip the proof here. And we also have the following bound, which has been shown in the proof of matrix completion if $\boldsymbol{\Delta}$ violates the condition (B.5).

$$\frac{1}{d_1 d_2} \|\boldsymbol{\Delta}\|_F^2 \leq \max\{C\alpha^2 \frac{r d' \log d'}{n}, C'\alpha^2 \frac{1}{n}\}. \tag{B.12}$$

Finally, we bound the term $\|\nabla \mathcal{L}_n(\mathbf{X}) - \nabla \bar{\mathcal{L}}_n(\mathbf{X})\|_2^2$. According to (B.11), we have

$$\nabla \bar{\mathcal{L}}_n(\mathbf{X}) = \frac{1}{p} \sum_{i=1}^n b_i \mathbf{A}_i,$$

$$\nabla \mathcal{L}_n(\mathbf{X}) = \frac{1}{p} \sum_{i=1}^n b_i' \mathbf{A}_i,$$

where we have

$$b_i = -\frac{f'(\langle \mathbf{A}_i, \mathbf{X} \rangle)}{f(\langle \mathbf{A}_i, \mathbf{X} \rangle)} f(\langle \mathbf{A}_i, \mathbf{X}^* \rangle) + \frac{f'(\langle \mathbf{A}_i, \mathbf{X} \rangle)}{1 - f(\langle \mathbf{A}_i, \mathbf{X} \rangle)} \big(1 - f(\langle \mathbf{A}_i, \mathbf{X}^* \rangle)\big),$$

$$b_i' = -\frac{f'(\langle \mathbf{A}_i, \mathbf{X} \rangle)}{f(\langle \mathbf{A}_i, \mathbf{X} \rangle)} \mathbb{1}_{(y_i=1)} + \frac{f'(\langle \mathbf{A}_i, \mathbf{X} \rangle)}{1 - f(\langle \mathbf{A}_i, \mathbf{X} \rangle)} \mathbb{1}_{(y_i=-1)}.$$

Therefore, we have

$$\nabla \mathcal{L}_n(\mathbf{X}) - \nabla \bar{\mathcal{L}}_n(\mathbf{X}) = \frac{1}{p} \sum_{i=1}^n (b_i' - b_i) \mathbf{A}_i,$$

which implies that

$$\|\nabla \mathcal{L}_n(\mathbf{X}) - \nabla \bar{\mathcal{L}}_n(\mathbf{X})\|_2 \leq \frac{1}{p} \Big\| \sum_{i=1}^n b_i' \mathbf{A}_i \Big\|_2 + \frac{1}{p} \Big\| \sum_{i=1}^n b_i \mathbf{A}_i \Big\|_2$$

$$\leq \frac{2}{p} \gamma_\alpha \Big\| \sum_{i=1}^n \mathbf{A}_i \Big\|_2.$$

Therefore, according to Lemma B.5, we have

$$\|\nabla \mathcal{L}_n(\mathbf{X}) - \nabla \bar{\mathcal{L}}_n(\mathbf{X})\|_2 \leq C \gamma_\alpha \sqrt{\frac{d_1 d_2 d' \log d'}{n}},$$

holds with probability at least $1 - C'/d$, which implies that

$$\|\nabla \mathcal{L}_n(\mathbf{X}) - \nabla \bar{\mathcal{L}}_n(\mathbf{X})\|_2^2 \leq C^2 \gamma_\alpha^2 \frac{d' \log d'}{p}. \tag{B.13}$$

Since the error bound (B.12) and (B.13) are dominated by the following statistical bound

$$C \max\{\gamma_\alpha^2, \alpha^2\} \frac{d' \log d'}{p},$$

where $C$ is the universal constant, we prove our results. $\qquad \square$

# C  Proofs of Technical Lemmas

In this section, we provide the proofs of several technical lemmas, which are used for proving our main theorem.

## C.1  Proof of Lemma A.1

In order to prove the local curvature condition, we need to make use of the following lemmas. Lemmas C.1 and C.2 are proved in Sections D.1 and D.2, respectively. In the following discussions, we denote $\widetilde{\mathbf{Z}} \in \mathbb{R}^{(d_1+d_2) \times r}$ as $\widetilde{\mathbf{Z}} = [\mathbf{U}; -\mathbf{V}]$. Recall $\mathbf{Z} = [\mathbf{U}; \mathbf{V}]$, then we have $\|\mathbf{U}^\top \mathbf{U} - \mathbf{V}^\top \mathbf{V}\|_F^2 = \|\widetilde{\mathbf{Z}}^\top \mathbf{Z}\|_F^2$, and $\nabla_{\mathbf{Z}}(\|\mathbf{U}^\top \mathbf{U} - \mathbf{V}^\top \mathbf{V}\|_F^2) = 4\widetilde{\mathbf{Z}}\widetilde{\mathbf{Z}}^\top \mathbf{Z}$.

**Lemma C.1** (Local Curvature Condition for Regularizer). Let $\mathbf{Z}, \mathbf{Z}^* \in \mathbb{R}^{(d_1+d_2) \times r}$. Denote the optimal rotation with respect to $\mathbf{Z}$ as $\mathbf{R} = \operatorname{argmin}_{\widetilde{\mathbf{R}} \in \mathbb{Q}_r} \|\mathbf{Z} - \mathbf{Z}^* \widetilde{\mathbf{R}}\|_F$, and $\mathbf{H} = \mathbf{Z} - \mathbf{Z}^* \mathbf{R}$. Consider the gradient of the regularization term $\|\widetilde{\mathbf{Z}}^\top \mathbf{Z}\|_F^2$, we have

$$\langle \widetilde{\mathbf{Z}}\widetilde{\mathbf{Z}}^\top \mathbf{Z}, \mathbf{H} \rangle \geq \frac{1}{2}\|\widetilde{\mathbf{Z}}^\top \mathbf{Z}\|_F^2 - \frac{1}{2}\|\widetilde{\mathbf{Z}}^\top \mathbf{Z}\|_F \cdot \|\mathbf{H}\|_F^2.$$

**Lemma C.2.** Suppose the expected loss function $\bar{\mathcal{L}}_n$ satisfies Conditions 5.5 and 5.6, then for any rank-$r$ matrices $\mathbf{X}, \mathbf{X}^* \in \mathbb{R}^{d_1 \times d_2}$, we have

$$\langle \nabla \bar{\mathcal{L}}_n(\mathbf{X}) - \nabla \bar{\mathcal{L}}_n(\mathbf{X}^*), \mathbf{X} - \mathbf{X}^* \rangle \geq \frac{1}{2\bar{L}}\|\nabla \bar{\mathcal{L}}_n(\mathbf{X}) - \nabla \bar{\mathcal{L}}_n(\mathbf{X}^*)\|_F^2 + \frac{\bar{\mu}}{2}\|\mathbf{X} - \mathbf{X}^*\|_F^2.$$

Besides, if we assume $\nabla \bar{\mathcal{L}}_n(\mathbf{X}^*) = 0$, then we have

$$\langle \nabla \bar{\mathcal{L}}_n(\mathbf{X}), \mathbf{X} - \mathbf{X}^* \rangle \geq \frac{1}{2\bar{L}}\|\nabla \bar{\mathcal{L}}_n(\mathbf{X})\|_F^2 + \frac{\bar{\mu}}{2}\|\mathbf{X} - \mathbf{X}^*\|_F^2.$$

Now, we are ready to prove Lemma A.1.

*Proof of Lemma A.1.* Recall that $\mathcal{L}_n$ and $\bar{\mathcal{L}}_n$ are the sample and expected loss functions, respectively. According to (A.1), we have

$$\langle \nabla \widetilde{F}_n(\mathbf{Z}), \mathbf{H} \rangle = \underbrace{\langle \nabla_{\mathbf{U}}\mathcal{L}_n(\mathbf{U}\mathbf{V}^\top), \mathbf{H}_U \rangle + \langle \nabla_{\mathbf{V}}\mathcal{L}_n(\mathbf{U}\mathbf{V}^\top), \mathbf{H}_V \rangle}_{I_1} + \frac{1}{2}\underbrace{\langle \widetilde{\mathbf{Z}}\widetilde{\mathbf{Z}}^\top \mathbf{Z}, \mathbf{H} \rangle}_{I_2}, \tag{C.1}$$

where $\widetilde{\mathbf{Z}} = [\mathbf{U}; -\mathbf{V}]$. Recall that $\mathbf{X}^* = \mathbf{U}^* \mathbf{V}^{*\top}$, and $\mathbf{X} = \mathbf{U}\mathbf{V}^\top$. Note that $\nabla_{\mathbf{U}}\mathcal{L}_n(\mathbf{U}\mathbf{V}^\top) = \nabla \mathcal{L}_n(\mathbf{X})\mathbf{V}$, and $\nabla_{\mathbf{V}}\mathcal{L}_n(\mathbf{U}\mathbf{V}^\top) = \nabla \mathcal{L}_n(\mathbf{X})^\top \mathbf{U}$. Thus, for the term $I_1$, we have

$$
\begin{aligned}
I_1 &= \langle \nabla \mathcal{L}_n(\mathbf{X}), \mathbf{U}\mathbf{V}^\top - \mathbf{U}^*\mathbf{V}^{*\top} + \mathbf{H}_U \mathbf{H}_V^\top \rangle \\
&= \underbrace{\langle \nabla \bar{\mathcal{L}}_n(\mathbf{X}), \mathbf{X} - \mathbf{X}^* + \mathbf{H}_U \mathbf{H}_V^\top \rangle}_{I_{11}} + \underbrace{\langle \nabla \mathcal{L}_n(\mathbf{X}) - \nabla \bar{\mathcal{L}}_n(\mathbf{X}), \mathbf{X} - \mathbf{X}^* + \mathbf{H}_U \mathbf{H}_V^\top \rangle}_{I_{12}}.
\end{aligned}
$$

First, we consider the term $I_{11}$. Recall that $\mathbf{X}^*$ minimizes the expected loss function $\bar{\mathcal{L}}_n(\mathbf{X})$, thus we have $\nabla \bar{\mathcal{L}}_n(\mathbf{X}^*) = 0$. By Lemma C.2, we have

$$\langle \nabla \bar{\mathcal{L}}_n(\mathbf{X}), \mathbf{X} - \mathbf{X}^* \rangle \geq \frac{\bar{\mu}}{2}\|\mathbf{X} - \mathbf{X}^*\|_F^2 + \frac{1}{2\bar{L}}\|\nabla \bar{\mathcal{L}}_n(\mathbf{X})\|_F^2. \tag{C.2}$$

Second, for the remaining term in $I_{11}$, we have

$$\left|\langle\nabla\bar{\mathcal{L}}_n(\mathbf{X}),\mathbf{H}_U\mathbf{H}_V^\top\rangle\right|\leq\|\nabla\bar{\mathcal{L}}_n(\mathbf{X})\|_F\cdot\|\mathbf{H}_U\mathbf{H}_V^\top\|_F\leq\frac{1}{2}\|\nabla\bar{\mathcal{L}}_n(\mathbf{X})\|_F\cdot\|\mathbf{H}\|_F^2,\qquad\text{(C.3)}$$

where the inequality holds because $|\langle\mathbf{A},\mathbf{B}\rangle|\leq\|\mathbf{A}\|_F\cdot\|\mathbf{B}\|_F$ and $2\|\mathbf{AB}\|_F\leq\|\mathbf{A}\|_F^2+\|\mathbf{B}\|_F^2$. Therefore, combining (C.2) and (C.3), the term $I_{11}$ can be lower bounded by

$$\begin{aligned}I_{11}&\geq\frac{\bar{\mu}}{2}\|\mathbf{X}-\mathbf{X}^*\|_F^2+\frac{1}{2\bar{L}}\|\nabla\bar{\mathcal{L}}_n(\mathbf{X})\|_F^2-\frac{1}{2}\|\nabla\bar{\mathcal{L}}_n(\mathbf{X})\|_F\cdot\|\mathbf{H}\|_F^2\\&\geq\frac{\bar{\mu}}{2}\|\mathbf{X}-\mathbf{X}^*\|_F^2+\frac{1}{4\bar{L}}\|\nabla\bar{\mathcal{L}}_n(\mathbf{X})\|_F^2-\frac{\bar{L}}{4}\|\mathbf{H}\|_F^4,\end{aligned}\qquad\text{(C.4)}$$

where the last inequality holds because $2ab\leq\beta a^2+b^2/\beta$, for any $\beta>0$. Next, for the term $I_{12}$, we have

$$\begin{aligned}\left|\langle\nabla\mathcal{L}_n(\mathbf{X})-\nabla\bar{\mathcal{L}}_n(\mathbf{X}),\mathbf{X}-\mathbf{X}^*\rangle\right|&\leq\|\nabla\mathcal{L}_n(\mathbf{X})-\nabla\bar{\mathcal{L}}_n(\mathbf{X})\|_2\cdot\|\mathbf{X}-\mathbf{X}^*\|_*\\&\leq\sqrt{2r}\|\nabla\mathcal{L}_n(\mathbf{X})-\nabla\bar{\mathcal{L}}_n(\mathbf{X}^*)\|_2\cdot\|\mathbf{X}-\mathbf{X}^*\|_F,\end{aligned}\qquad\text{(C.5)}$$

where the first inequality follows from the Von Neumann trace inequality, and the second inequality follows from the fact that $\mathrm{rank}(\mathbf{X}-\mathbf{X}^*)\leq2r$. For the remaining term in $I_{12}$, by similar techniques, we have

$$\left|\langle\nabla\mathcal{L}_n(\mathbf{X})-\nabla\bar{\mathcal{L}}_n(\mathbf{X}),\mathbf{H}_U\mathbf{H}_V^\top\rangle\right|\leq\sqrt{2r}\|\nabla\mathcal{L}_n(\mathbf{X})-\nabla\bar{\mathcal{L}}_n(\mathbf{X})\|_2\cdot\|\mathbf{H}_U\mathbf{H}_V^\top\|_F.\qquad\text{(C.6)}$$

Thus, combining (C.5) and (C.6), the term $I_{12}$ can be lower bounded by

$$\begin{aligned}I_{12}&\geq-\sqrt{2r}\|\nabla\mathcal{L}_n(\mathbf{X})-\nabla\bar{\mathcal{L}}_n(\mathbf{X})\|_2\cdot\left(\|\mathbf{X}-\mathbf{X}^*\|_F+\frac{1}{2}\|\mathbf{H}\|_F^2\right)\\&\geq-\frac{\bar{\mu}}{4}\|\mathbf{X}-\mathbf{X}^*\|_F^2-\frac{\bar{L}}{4}\|\mathbf{H}\|_F^4-\left(\frac{r}{2\bar{L}}+\frac{2r}{\bar{\mu}}\right)\|\nabla\mathcal{L}_n(\mathbf{X})-\nabla\bar{\mathcal{L}}_n(\mathbf{X})\|_2^2,\end{aligned}\qquad\text{(C.7)}$$

where the first inequality holds because $2\|\mathbf{AB}\|_F\leq\|\mathbf{A}\|_F^2+\|\mathbf{B}\|_F^2$, and the last inequality follows from the fact that $2ab\leq\beta a^2+b^2/\beta$, for any $\beta>0$. Therefore, combining (C.4) and (C.7), we obtain the lower bound of $I_1$

$$I_1\geq\frac{\bar{\mu}}{4}\|\mathbf{X}-\mathbf{X}^*\|_F^2+\frac{1}{4\bar{L}}\|\nabla\bar{\mathcal{L}}_n(\mathbf{X})\|_F^2-\frac{\bar{L}}{2}\|\mathbf{H}\|_F^4-\left(\frac{r}{2\bar{L}}+\frac{2r}{\bar{\mu}}\right)\|\nabla\mathcal{L}_n(\mathbf{X})-\nabla\bar{\mathcal{L}}_n(\mathbf{X})\|_2^2.\qquad\text{(C.8)}$$

On the other hand, for the term $I_2$, according to lemma C.1, we have

$$I_2\geq\frac{1}{2}\|\widetilde{\mathbf{Z}}^\top\mathbf{Z}\|_F^2-\frac{1}{2}\|\widetilde{\mathbf{Z}}^\top\mathbf{Z}\|_F\cdot\|\mathbf{H}\|_F^2\geq\frac{1}{4}\|\widetilde{\mathbf{Z}}^\top\mathbf{Z}\|_F^2-\frac{1}{4}\|\mathbf{H}\|_F^4,\qquad\text{(C.9)}$$

where the last inequality holds because $2ab\leq a^2+b^2$. By plugging (C.8) and (C.9) into (C.1), we have

$$\begin{aligned}\langle\nabla\widetilde{F}_n(\mathbf{Z}),\mathbf{H}\rangle&\geq\frac{\bar{\mu}}{4}\|\mathbf{X}-\mathbf{X}^*\|_F^2+\frac{1}{4\bar{L}}\|\nabla\bar{\mathcal{L}}_n(\mathbf{X})\|_F^2+\frac{1}{8}\|\widetilde{\mathbf{Z}}^\top\mathbf{Z}\|_F^2\\&\quad-\frac{4\bar{L}+1}{8}\|\mathbf{H}\|_F^4-\left(\frac{r}{2\bar{L}}+\frac{2r}{\bar{\mu}}\right)\|\nabla\mathcal{L}_n(\mathbf{X})-\nabla\bar{\mathcal{L}}_n(\mathbf{X})\|_2^2.\end{aligned}\qquad\text{(C.10)}$$

Furthermore, denote $\widetilde{\mathbf{Z}}^* = [\mathbf{U}^*; -\mathbf{V}^*]$, then we have

$$
\begin{aligned}
\|\widetilde{\mathbf{Z}}^\top \mathbf{Z}\|_F^2 &= \langle \mathbf{Z}\mathbf{Z}^\top - \mathbf{Z}^*\mathbf{Z}^{*\top}, \widetilde{\mathbf{Z}}\widetilde{\mathbf{Z}}^\top - \widetilde{\mathbf{Z}}^*\widetilde{\mathbf{Z}}^{*\top} \rangle + \langle \mathbf{Z}^*\mathbf{Z}^{*\top}, \widetilde{\mathbf{Z}}\widetilde{\mathbf{Z}}^\top \rangle + \langle \mathbf{Z}\mathbf{Z}^\top, \widetilde{\mathbf{Z}}^*\widetilde{\mathbf{Z}}^{*\top} \rangle \\
&\geq \langle \mathbf{Z}\mathbf{Z}^\top - \mathbf{Z}^*\mathbf{Z}^{*\top}, \widetilde{\mathbf{Z}}\widetilde{\mathbf{Z}}^\top - \widetilde{\mathbf{Z}}^*\widetilde{\mathbf{Z}}^{*\top} \rangle \\
&= \|\mathbf{U}\mathbf{U}^\top - \mathbf{U}^*\mathbf{U}^{*\top}\|_F^2 + \|\mathbf{V}\mathbf{V}^\top - \mathbf{V}^*\mathbf{V}^{*\top}\|_F^2 - 2\|\mathbf{U}\mathbf{V}^\top - \mathbf{U}^*\mathbf{V}^{*\top}\|_F^2, \quad \text{(C.11)}
\end{aligned}
$$

where the first equality holds because $\widetilde{\mathbf{Z}}^{*\top}\mathbf{Z}^* = 0$, and the inequality holds because $\langle \mathbf{A}\mathbf{A}^\top, \mathbf{B}\mathbf{B}^\top \rangle = \|\mathbf{A}^\top \mathbf{B}\|_F^2 \geq 0$. Thus, according to Lemma E.2, we have

$$
4\|\mathbf{X} - \mathbf{X}^*\|_F^2 + \|\widetilde{\mathbf{Z}}^\top \mathbf{Z}\|_F^2 = \|\mathbf{Z}\mathbf{Z}^\top - \mathbf{Z}^*\mathbf{Z}^{*\top}\|_F^2 \geq 4(\sqrt{2}-1)\sigma_r\|\mathbf{H}\|_F^2, \quad \text{(C.12)}
$$

where the first inequality follows from (C.11), and the second inequality follows from Lemma E.2 and the fact that $\sigma_r^2(\mathbf{Z}^*) = 2\sigma_r$. Denote $\bar{\mu}' = \min\{\bar{\mu}, 1\}$. Therefore, plugging (C.12) into (C.10), we have

$$
\begin{aligned}
\langle \nabla \widetilde{F}_n(\mathbf{Z}), \mathbf{H} \rangle &\geq \frac{\bar{\mu}'\sigma_r}{10}\|\mathbf{H}\|_F^2 + \frac{1}{4\bar{L}}\|\nabla\bar{\mathcal{L}}_n(\mathbf{X})\|_F^2 + \frac{1}{16}\|\widetilde{\mathbf{Z}}^\top \mathbf{Z}\|_F^2 \\
&\quad - \frac{4\bar{L}+1}{8}\|\mathbf{H}\|_F^4 - \left(\frac{r}{2\bar{L}} + \frac{2r}{\bar{\mu}}\right)\|\nabla\mathcal{L}_n(\mathbf{X}) - \nabla\bar{\mathcal{L}}_n(\mathbf{X})\|_2^2,
\end{aligned}
$$

which completes the proof. $\qquad\square$

## C.2 Proof of Lemma A.2

*Proof.* According to (A.1), for the term $\|\nabla \widetilde{F}_n(\mathbf{Z})\|_F^2$, we have

$$
\begin{aligned}
\|\nabla \widetilde{F}_n(\mathbf{Z})\|_F^2 &= \|\nabla_{\mathbf{U}}\mathcal{L}_n(\mathbf{U}\mathbf{V}^\top) + \tfrac{1}{2}\mathbf{U}(\mathbf{U}^\top\mathbf{U} - \mathbf{V}^\top\mathbf{V})\|_F^2 + \|\nabla_{\mathbf{V}}\mathcal{L}_n(\mathbf{U}\mathbf{V}^\top) + \tfrac{1}{2}\mathbf{V}(\mathbf{V}^\top\mathbf{V} - \mathbf{U}^\top\mathbf{U})\|_F^2 \\
&\leq 2\|\nabla_{\mathbf{U}}\mathcal{L}_n(\mathbf{U}\mathbf{V}^\top)\|_F^2 + 2\|\nabla_{\mathbf{V}}\mathcal{L}_n(\mathbf{U}\mathbf{V}^\top)\|_F^2 + \tfrac{1}{2}\|\mathbf{U}^\top\mathbf{U} - \mathbf{V}^\top\mathbf{V}\|_F^2 \cdot (\|\mathbf{U}\|_2^2 + \|\mathbf{V}\|_2^2) \\
&\leq 2\|\nabla_{\mathbf{U}}\mathcal{L}_n(\mathbf{U}\mathbf{V}^\top)\|_F^2 + 2\|\nabla_{\mathbf{V}}\mathcal{L}_n(\mathbf{U}\mathbf{V}^\top)\|_F^2 + \|\mathbf{U}^\top\mathbf{U} - \mathbf{V}^\top\mathbf{V}\|_F^2 \cdot \|\mathbf{Z}\|_2^2, \quad \text{(C.13)}
\end{aligned}
$$

where the first inequality holds because $\|\mathbf{A} + \mathbf{B}\|_F^2 \leq 2\|\mathbf{A}\|_F^2 + 2\|\mathbf{B}\|_F^2$ and $\|\mathbf{A}\mathbf{B}\|_F \leq \|\mathbf{A}\|_2 \cdot \|\mathbf{B}\|_F$, and the last inequality holds because $\max\{\|\mathbf{U}\|_2, \|\mathbf{V}\|_2\} \leq \|\mathbf{Z}\|_2$. Furthermore, we have

$$
\begin{aligned}
\|\nabla_{\mathbf{U}}\mathcal{L}_n(\mathbf{U}\mathbf{V}^\top)\|_F^2 &= \|\nabla\mathcal{L}_n(\mathbf{X})\mathbf{V}\|_F^2 \\
&\leq 2\|\big(\nabla\mathcal{L}_n(\mathbf{X}) - \nabla\bar{\mathcal{L}}_n(\mathbf{X})\big)\mathbf{V}\|_F^2 + 2\|\nabla\bar{\mathcal{L}}_n(\mathbf{X})\mathbf{V}\|_F^2 \\
&\leq 2r\|\nabla\mathcal{L}_n(\mathbf{X}) - \nabla\bar{\mathcal{L}}_n(\mathbf{X})\|_2^2 \cdot \|\mathbf{V}\|_2^2 + 2\|\nabla\bar{\mathcal{L}}_n(\mathbf{X})\|_F^2 \cdot \|\mathbf{V}\|_2^2, \quad \text{(C.14)}
\end{aligned}
$$

where the first inequality follows from the triangle inequality, and the last inequality holds because $\text{rank}(\mathbf{V}) = r$, and $\|\mathbf{A}\mathbf{B}\|_F \leq \|\mathbf{A}\|_2 \cdot \|\mathbf{B}\|_F$. Similarly, we have

$$
\begin{aligned}
\|\nabla_{\mathbf{V}}\mathcal{L}_n(\mathbf{U}\mathbf{V}^\top)\|_F^2 &= \|\nabla\mathcal{L}_n(\mathbf{X})^\top\mathbf{U}\|_F^2 \\
&\leq 2\|\big(\nabla\mathcal{L}_n(\mathbf{X}) - \nabla\bar{\mathcal{L}}_n(\mathbf{X})\big)^\top\mathbf{U}\|_F^2 + 2\|\nabla\bar{\mathcal{L}}_n(\mathbf{X})^\top\mathbf{U}\|_F^2 \\
&\leq 2r\|\nabla\mathcal{L}_n(\mathbf{X}) - \nabla\bar{\mathcal{L}}_n(\mathbf{X})\|_2^2 \cdot \|\mathbf{U}\|_2^2 + 2\|\nabla\bar{\mathcal{L}}_n(\mathbf{X})\|_F^2 \cdot \|\mathbf{U}\|_2^2. \quad \text{(C.15)}
\end{aligned}
$$

Therefore, by plugging (C.14) and (C.15) into (C.13), we obtain

$$
\|\nabla\widetilde{F}_n(\mathbf{Z})\|_F^2 \leq \big(8\|\nabla\bar{\mathcal{L}}_n(\mathbf{X})\|_F^2 + \|\mathbf{U}^\top\mathbf{U} - \mathbf{V}^\top\mathbf{V}\|_F^2 + 8r\|\nabla\mathcal{L}_n(\mathbf{X}) - \nabla\bar{\mathcal{L}}_n(\mathbf{X})\|_2^2\big) \cdot \|\mathbf{Z}\|_2^2,
$$

where the inequality holds because $\max\{\|\mathbf{U}\|_2, \|\mathbf{V}\|_2\} \leq \|\mathbf{Z}\|_2$. Thus, we complete the proof. $\qquad\square$

# D Proofs of Auxiliary Lemmas in Appendix C

In this section, we prove the auxiliary lemmas used in Appendix C.

## D.1 Proof of Lemma C.1

*Proof.* Recall $\mathbf{H} = \mathbf{Z} - \mathbf{Z}^* \mathbf{R}$, then we have

$$
\begin{aligned}
\langle \widetilde{\mathbf{Z}} \widetilde{\mathbf{Z}}^\top \mathbf{Z}, \mathbf{H} \rangle &= \frac{1}{2} \langle \widetilde{\mathbf{Z}}^\top \mathbf{Z}, \widetilde{\mathbf{Z}}^\top \mathbf{Z} \rangle + \frac{1}{2} \langle \widetilde{\mathbf{Z}}^\top \mathbf{Z}, \widetilde{\mathbf{Z}}^\top \mathbf{Z} - 2 \widetilde{\mathbf{Z}}^\top \mathbf{Z}^* \mathbf{R} \rangle \\
&= \frac{1}{2} \| \widetilde{\mathbf{Z}}^\top \mathbf{Z} \|_F^2 + \frac{1}{2} \langle \widetilde{\mathbf{Z}}^\top \mathbf{Z}, \widetilde{\mathbf{Z}}^\top \mathbf{H} - \widetilde{\mathbf{Z}}^\top \mathbf{Z}^* \mathbf{R} \rangle.
\end{aligned} \tag{D.1}
$$

Denote $\widetilde{\mathbf{Z}}^* = [\mathbf{U}^*; -\mathbf{V}^*]$, we have $\mathbf{Z}^{*\top} \widetilde{\mathbf{Z}} = \widetilde{\mathbf{Z}}^{*\top} \mathbf{Z}$. Since $\widetilde{\mathbf{Z}}^\top \mathbf{Z}$ is symmetric, thus we have

$$
\begin{aligned}
\langle \widetilde{\mathbf{Z}}^\top \mathbf{Z}, \widetilde{\mathbf{Z}}^\top \mathbf{Z}^* \mathbf{R} \rangle &= \langle \widetilde{\mathbf{Z}}^\top \mathbf{Z}, \mathbf{R}^\top \mathbf{Z}^{*\top} \widetilde{\mathbf{Z}} \rangle \\
&= \langle \widetilde{\mathbf{Z}}^\top \mathbf{Z}, \mathbf{R}^\top \widetilde{\mathbf{Z}}^{*\top} \mathbf{Z} \rangle \\
&= \langle \widetilde{\mathbf{Z}}^\top \mathbf{Z}, \mathbf{R}^\top \widetilde{\mathbf{Z}}^{*\top} \mathbf{H} \rangle,
\end{aligned} \tag{D.2}
$$

where the last inequality holds because $\widetilde{\mathbf{Z}}^{*\top} \mathbf{Z}^* = 0$. Therefore, by plugging (D.2) into (D.1), we have

$$
\begin{aligned}
| \langle \widetilde{\mathbf{Z}}^\top \mathbf{Z}, \widetilde{\mathbf{Z}}^\top \mathbf{H} - \widetilde{\mathbf{Z}}^\top \mathbf{Z}^* \mathbf{R} \rangle | &= | \langle \widetilde{\mathbf{Z}}^\top \mathbf{Z}, (\widetilde{\mathbf{Z}} - \widetilde{\mathbf{Z}}^* \mathbf{R})^\top \mathbf{H} \rangle | \\
&\leq \| \widetilde{\mathbf{Z}}^\top \mathbf{Z} \|_F \cdot \| \widetilde{\mathbf{Z}} - \widetilde{\mathbf{Z}}^* \mathbf{R} \|_F \cdot \| \mathbf{H} \|_F \\
&\leq \| \widetilde{\mathbf{Z}}^\top \mathbf{Z} \|_F \cdot \| \mathbf{H} \|_F^2,
\end{aligned} \tag{D.3}
$$

where the first inequality holds because $| \langle \mathbf{A}, \mathbf{BC} \rangle | \leq \| \mathbf{A} \|_F \cdot \| \mathbf{BC} \|_F \leq \| \mathbf{A} \|_F \cdot \| \mathbf{B} \|_F \cdot \| \mathbf{C} \|_F$, and the second inequality holds because $\| \widetilde{\mathbf{Z}} - \widetilde{\mathbf{Z}}^* \mathbf{R} \|_F = \| \mathbf{Z} - \mathbf{Z}^* \mathbf{R} \|_F = \| \mathbf{H} \|_F$. Therefore, according to (D.3), we have

$$
\langle \widetilde{\mathbf{Z}} \widetilde{\mathbf{Z}}^\top \mathbf{Z}, \mathbf{H} \rangle \geq \frac{1}{2} \| \widetilde{\mathbf{Z}}^\top \mathbf{Z} \|_F^2 - \frac{1}{2} \| \widetilde{\mathbf{Z}}^\top \mathbf{Z} \|_F \cdot \| \mathbf{H} \|_F^2,
$$

which completes the proof. $\qquad \square$

## D.2 Proof of Lemma C.2

*Proof.* According to the restricted strongly convexity Condition 5.5, we have

$$
\bar{\mathcal{L}}_n(\mathbf{X}^*) \geq \bar{\mathcal{L}}_n(\mathbf{X}) + \langle \nabla \bar{\mathcal{L}}_n(\mathbf{X}), \mathbf{X}^* - \mathbf{X} \rangle + \frac{\bar{\mu}}{2} \| \mathbf{X} - \mathbf{X}^* \|_F^2. \tag{D.4}
$$

Besides, according to lemma E.4, we have

$$
\bar{\mathcal{L}}_n(\mathbf{X}) - \bar{\mathcal{L}}_n(\mathbf{X}^*) \geq \langle \nabla \bar{\mathcal{L}}_n(\mathbf{X}^*), \mathbf{X} - \mathbf{X}^* \rangle + \frac{1}{2\bar{L}} \| \nabla \bar{\mathcal{L}}_n(\mathbf{X}) - \nabla \bar{\mathcal{L}}_n(\mathbf{X}^*) \|_F^2. \tag{D.5}
$$

Therefore, combining (D.5) and (D.4), we have

$$
\langle \nabla \bar{\mathcal{L}}_n(\mathbf{X}) - \nabla \bar{\mathcal{L}}_n(\mathbf{X}^*), \mathbf{X} - \mathbf{X}^* \rangle \geq \frac{1}{2\bar{L}} \| \nabla \bar{\mathcal{L}}_n(\mathbf{X}) - \nabla \bar{\mathcal{L}}_n(\mathbf{X}^*) \|_F^2 + \frac{\bar{\mu}}{2} \| \mathbf{X} - \mathbf{X}^* \|_F^2,
$$

which completes the proof. $\qquad \square$

# E   Other Auxiliary lemmas

**Lemma E.1.** (Tu et al., 2015) Let $\mathbf{M}_1, \mathbf{M}_2 \in \mathbb{R}^{d_1 \times d_2}$ be two rank $r$ matrices. Suppose they have SVDs $\mathbf{M}_1 = \mathbf{U}_1 \boldsymbol{\Sigma}_1 \mathbf{V}_1^\top$ and $\mathbf{M}_2 = \mathbf{U}_2 \boldsymbol{\Sigma}_2 \mathbf{V}_2^\top$. Suppose $\|\mathbf{M}_1 - \mathbf{M}_2\|_2 \leq \sigma_r(\mathbf{M}_1)/2$, then we have

$$d^2\left([\mathbf{U}_2; \mathbf{V}_2]\boldsymbol{\Sigma}_2^{1/2}, [\mathbf{U}_1; \mathbf{V}_1]\boldsymbol{\Sigma}_2^{1/2}\right) \leq \frac{2}{\sqrt{2}-1} \frac{\|\mathbf{M}_2 - \mathbf{M}_1\|_F^2}{\sigma_r(\mathbf{M}_1)}.$$

**Lemma E.2.** (Tu et al., 2015) For any $\mathbf{Z}, \mathbf{Z}' \in \mathbb{R}^{(d_1+d_2) \times r}$, we have

$$d^2(\mathbf{Z}, \mathbf{Z}') \leq \frac{1}{2(\sqrt{2}-1)\sigma_r^2(\mathbf{Z}')} \|\mathbf{Z}\mathbf{Z}^\top - \mathbf{Z}'\mathbf{Z}'^\top\|_F^2.$$

**Lemma E.3.** (Tu et al., 2015) For any $\mathbf{Z}, \mathbf{Z}' \in \mathbb{R}^{(d_1+d_2) \times r}$ satisfying $d(\mathbf{Z}, \mathbf{Z}') \leq \|\mathbf{Z}'\|_2/4$, we have

$$\|\mathbf{Z}\mathbf{Z}^\top - \mathbf{Z}'\mathbf{Z}'^\top\|_F \leq \frac{9}{4}\|\mathbf{Z}'\|_2 \cdot d(\mathbf{Z}, \mathbf{Z}').$$

**Lemma E.4.** (Nesterov, 2004) Let $\mathcal{L} : \mathbb{R}^{d_1 \times d_2} \to \mathbb{R}$ be convex and $L$-smooth, then for any $\mathbf{X}, \mathbf{Y} \in \mathbb{R}^{d_1 \times d_2}$, we have

$$\mathcal{L}(\mathbf{Y}) \geq \mathcal{L}(\mathbf{X}) + \langle \nabla \mathcal{L}(\mathbf{X}), \mathbf{Y} - \mathbf{X} \rangle + \frac{1}{2L} \|\nabla \mathcal{L}(\mathbf{Y}) - \nabla \mathcal{L}(\mathbf{X})\|_F^2.$$